\documentclass{article}

\PassOptionsToPackage{dvipsnames}{xcolor}
\usepackage{iclr2019_conference,times}

\usepackage[utf8]{inputenc} % allow utf-8 input
\usepackage[T1]{fontenc}    % use 8-bit T1 fonts
\usepackage{booktabs}       % professional-quality tables
\usepackage{multirow}
\usepackage{amsfonts}       % blackboard math symbols
\usepackage{nicefrac}       % compact symbols for 1/2, etc.
\usepackage{microtype}      % microtypography
\usepackage{amsmath}
\usepackage{algpseudocode,algorithm,algorithmicx}
\usepackage{xspace}
\usepackage{pifont}
\usepackage{graphicx}
\usepackage{wrapfig}
\graphicspath{{figures/}}
\usepackage{lipsum}
\usepackage{caption}
\usepackage{enumerate}

\newcommand{\cmark}{{\color{ForestGreen}\ding{51}}}
\newcommand{\xmark}{{\color{BrickRed}\ding{55}}}
\usepackage{url}            % simple URL typesetting
\usepackage{hyperref}       % hyperlinks
\hypersetup{breaklinks=true,colorlinks}

\newcommand{\SF}[1]{{\color{BrickRed}{[@Sanja: #1]}}}

\def\PMN{PMN}
\def\E{\mathcal{E}}
\def\Mobj{M_\mathrm{obj}}
\def\Matt{M_\mathrm{att}}
\def\Mrel{M_\mathrm{rel}}
\def\Mcnt{M_\mathrm{cnt}}
\def\Mcap{M_\mathrm{cap}}
\def\Mvqa{M_\mathrm{vqa}}
\newcommand{\rMrel}[1]{M_{\mathrm{rel}_{#1}}}
\newcommand{\rMcnt}[1]{M_{\mathrm{cnt}_{#1}}}
\newcommand{\rMcap}[1]{M_{\mathrm{cap}_{#1}}}
\newcommand{\rMvqa}[1]{M_{\mathrm{vqa}_{#1}}}

\makeatletter
\DeclareRobustCommand\onedot{\futurelet\@let@token\@onedot}
\def\@onedot{\ifx\@let@token.\else.\null\fi\xspace}

\def\eg{\emph{e.g}\onedot} 
\def\ie{\emph{i.e}\onedot} 
\def\cf{\emph{c.f}\onedot}

\makeatother

\title{Visual Reasoning by \\ Progressive Module Networks}
\author{
  Seung Wook Kim$^{1,2}$ , Makarand Tapaswi$^{1,2}$ , Sanja Fidler$^{1,2,3}$  \\
  $^1$Department of Computer Science, University of Toronto\\
  $^2$Vector Institute, Canada\\
  $^3$NVIDIA\\
  \texttt{\{seung,makarand,fidler\}@cs.toronto.edu} \\
}
\begin{document}

\maketitle
\vspace{-1.5mm}
%%%%%%%%%%%%%%%%%%%%%%%%%%%%%%%%%%%%%%%%%%%%%%%%%%%%%%%%%%%%%%%%%%%%%%%%%%%%
\begin{abstract}
Humans learn to solve tasks of increasing complexity by building on top of previously acquired knowledge. % and broadening their skill set.
Typically, there exists a natural progression in the tasks that we learn -- most do not require completely independent solutions, but can be broken down into simpler subtasks.
We propose to represent a solver for each task as a neural module that calls existing modules (solvers for simpler tasks) in a functional program-like manner.
Lower modules are a black box to the calling module, and communicate only via a query and an output.
Thus, a module for a new task learns to query existing modules and composes their outputs in order to produce its own output.
%Each module also contains a \emph{residual} component that learns to solve aspects of the new task that lower modules cannot solve.
Our model effectively combines previous skill-sets, does not suffer from forgetting, and is fully differentiable.
% It is also more interpretable, since it makes the reasoning process explicit.
We test our model in learning a set of visual reasoning tasks, and demonstrate improved performances in all tasks by learning progressively. %state-of-the-art performance in Visual Question Answering, the highest-level task in our task set.
By evaluating the reasoning process using human judges, we show that our model is more interpretable than an attention-based baseline.

%The ability to progressively improve skills and broaden knowledge is a fundamental characteristic of intelligence.
%Humans are a prime example as we progressively acquire new knowledge over time, and can perform multiple tasks with different levels of complexity with the learned knowledge.
%A high level task that requires complex reasoning can often be decomposed into a number of simpler low level tasks where the process of decomposition can be applied recursively.

%Motivated by this, we propose Progressive Module Networks (PMN) that constructs modules for solving specific tasks based on current existing task modules.
%We show the effectiveness of PMN by building modules for several computer vision tasks.
%In particular, we obtain {\color{gray}state-of-the-art} performance on the meta-task of Visual Question Answering, that combines information from several modules.
\end{abstract}

%%%%%%%%%%%%%%%%%%%%%%%%%%%%%%%%%%%%%%%%%%%%%%%%%%%%%%%%%%%%%%%%%%%%%%%%%%%%

%!TEX root = paper.tex
\vspace{-2.5mm}
\section{Introduction}
\label{sec:intro}
\vspace{-1.0mm}

Humans acquire skills and knowledge in a curriculum by building on top of previously acquired knowledge.
For example, in school we first learn simple mathematical operations such as addition and multiplication before moving on to solving equations.
Similarly, the ability to %recognize individual objects is a pre-requisite for counting them; and
answer complex visual questions often requires the skills to understand attributes such as color, recognize a variety of objects, and be able to spatially relate them.
%Just as humans, we show that machines can also benefit by learning tasks in progressive complexity sequentially and composing knowledge along the way.
Just like humans, machines may also benefit by sequentially learning tasks in progressive complexity and composing knowledge along the way.

%An intelligent machine that mimics human behavior should be able to perform multiple tasks that require different levels of complexity.
%Consider a situation where a robot is asked to describe an image.
%The robot has to have an ability to do simpler tasks such as locating objects in the image, identifying the names  of the objects, and inferring relationships between the objects.
%It would be also helpful if it could answer questions related to objects or activities in the image.
%Currently, most state of the art models implemented as neural networks are good at solving one specific task.
%However, as tasks become more complicated, there will be a need to progressively learn by making use of previously learned knowledge just as humans do.

The process of training a machine learning model to be able to solve multiple tasks, or \textit{multi-task learning} (MTL), has been widely studied~\citep{long17mtl_reln,ruder17overview_mtl,ruder17sluice,rusu16}.
The dominant approach is to have a model that shares parameters (\eg, bottom layers of a CNN) with individualized prediction heads~\citep{caruana1993,long17mtl_reln}.
By sharing parameters, models are able to learn better task-agnostic data representations.
However, the tasks are disconnected as their outputs are not combined to solve tasks of increasing complexity.
It is desirable if one task can learn to process the predictions from other tasks thereby reaping the benefits of MTL.

%We address the problem of MTL where tasks naturally progress in complexity. %, and their solutions build on top of each other.
%In \citet{andreas16}, Neural Module Networks (NMN) that call a sequence of \emph{modules} to solve a complex task are introduced.
% Each module is designed to solve a simpler subtask.
%By utilizing modules with architectures tailored to solve a subproblem, they showed improved performance on the higher-level task.
%In particular, \citet{andreas16,hu17}~addressed Visual Question Answering (VQA), where, given a question, a  sequence of modules such as \emph{describe-region} or \emph{find-object} are called.
%This sequence was either parsed from the question~\citep{andreas16}, or required training via policy gradient optimization~\citep{hu17}.

%Neural Module Networks (NMN)~\cite{andreas16,hu17} addresses visual question answering where \emph{questions} have compositional structure.
%Their modules are simple modules that solve basic problems such as \emph{describe-region} or \emph{find-object}.
%We make each \emph{module} compositional such that a module execution is composed of calling other modules internally.
%It can be thought of as a comptuer program making use of available libraries without having to know the internal operations.
%By having this module level compositionality, we can build more and more complex task solvers from simpler task solvers.

In this paper, we address the problem of MTL where tasks exhibit a natural progression in complexity.
We propose Progressive Module Networks (PMN), a framework for multi-task learning by progressively designing modules on top of existing modules.
Each module is a neural network that can query modules for lower-level tasks, which in turn may query modules for even simpler tasks.
The modules communicate by learning to query other modules and process their outputs, while the internal module processes are a blackbox.
This is similar to a computer program that uses available libraries without having to know their internal operations.
Parent modules can choose which lower-level modules they want to query via a soft gating mechanism. %, and let them learn to produce queries for those modules.
%Additionally, each module can optionally have a ``residual'' submodule that learns to address aspects of the new task that simpler modules cannot.
Examining the queries, replies, and choices a parent module makes, we can understand the reasoning behind the module's output.

%PMN can be seen as a generalization of Neural Module Networks (NMN)~\citep{andreas16,hu17} in the sense that PMN's modules are task-level modules.
PMN's modular structure is related to ~\citet{andreas16} and ~\citet{hu17}, but PMN's modules are task-level modules.
PMN is compositional, \ie~modules build on modules which build on modules, and is fully differentiable.
It allows efficient use of data by not needing to re-learn previously acquired knowledge. % again for a new task.
By learning selective information flow between modules, interpretability arises naturally.
% It also provides a useful way to understand model behavior.
%Neural networks are regarded as black boxes because their inner workings are hard to analyze.

We demonstrate PMN in learning a set of visual reasoning tasks such as counting, captioning, and Visual Question Answering (VQA).
PMN outperform baselines without module composition on all tasks.
We further analyze the interpretability of PMN's reasoning process with human judges.

\vspace{-2.0mm}
\section{Related Work}
\label{sec:relwork}
\vspace{-1.5mm}

\textbf{Multi-task learning.}
%There exists a vast amount of literature on MTL~\cite{ruder17overview_mtl}.
%Here, we review work closest to ours.
The dominant approach to multi-task learning is to have a model that shares parameters in a soft~\citep{duong15softmtl,yang17softmtl} or hard way~\citep{caruana1993}.
Soft sharing refers to each task having independent weights that are constrained to be similar (\eg~via $\ell_2$ regularization~\citep{duong15softmtl}, trace norm~\citep{yang17softmtl}) while hard sharing typically means that all tasks share the base network but have independent layers on the top ~\citep{kokkinos17ubernet,misra16cross_stitch}.
While sharing parameters helps to compute a task-agnostic representation that is not overfit to a specific task, tasks do not directly share information or help each other.
%It would be more desirable if one task can directly use information learned by other tasks to maximally enjoy the benefit of multi-task learning.

\citet{bilen16} propose the Multinet architecture where tasks can interact with each other in addition to shared image features.
Multinet solves one task at each time step and appends the encoded output of each task to existing data representation. % starting from image features from a CNN.
%Thus, at the next time step, the new task uses enriched data representation.
A similar idea, Progressive Neural Networks (PNNs)~\citep{rusu16} use a new neural network for each task, but are designed to prevent catastrophic forgetting as they transfer knowledge from previous tasks by making lateral connections to representations of previously learned tasks. % instantiating a new neural network for each task being solved while
%In both Multinet and PNN, multiple tasks interact with each other in an indirect fashion as they are mainly used to learn a better data representation.
%We go one step further, by enabling task-wise interactions.
Recently, \citet{wang2017vqa} propose the VQA-Machine which exploits a set of existing algorithms to solve the VQA problem.
\citet{zamir2018taskonomy} learn a computational taxonomy for task transfer learning on several vision problems.
However, the major differences to this work are PMN's compositional modular structure, ability to directly query other modules, and the overall process of learning increasingly complex tasks.

\iffalse
Parent modules in PMN choose and call lower modules, and use the outputs of queried modules directly to solve their task.
This interaction lets PMN be more data efficient as modules can be trained on different datasets and the knowledge utilized by other tasks. %knowledge learned in each task or dataset can be utilized by other tasks.
PMNs are also a stepping stone to train a universal and intelligent machine that progressively improves itself by learning new tasks, while relating them to the ones it already knows.
\fi

\textbf{Module networks.}
%There has been interest in modular architecture in neural networks.
Pioneering work in modular structure, NMN~\citep{andreas16,hu17} addresses VQA where questions have a compositional structure.
Given an inventory of small networks, or modules, NMN produces a layout for assembling the modules for any question.
We extend their modularity idea further and treat \emph{each task} as compositional.
PMN is more general and can be used for any arbitrary task where there exists an exploitable progressive learning sequence.

\textbf{Visual question answering.}
VQA has seen great progress in recent years: improved multimodal pooling functions~\citep{fukui16mcb, kim2018bilinear}, multi-hop attention~\citep{yang16san}, driving attention through both bottom-up and top-down schemes~\citep{anderson17}, and modeling attention between words and image regions recurrently~\citep{hudson2018compositional} are some of the important advances.
There are also attempts to generate programs or sequence of modules automatically that yield a list of interpretable steps~\citep{hu17,johnson17clevrprogram} using policy gradient optimization.
Our approach treats visual reasoning as a compositional multi-task problem, and shows that using sub-tasks compositionally can help improve performance and interpretability.

%!TEX root = paper.tex
\begin{figure}[t!]
\vspace{-10mm}
\begin{minipage}{0.56\linewidth}
\begin{center}
\includegraphics[width=1\columnwidth]{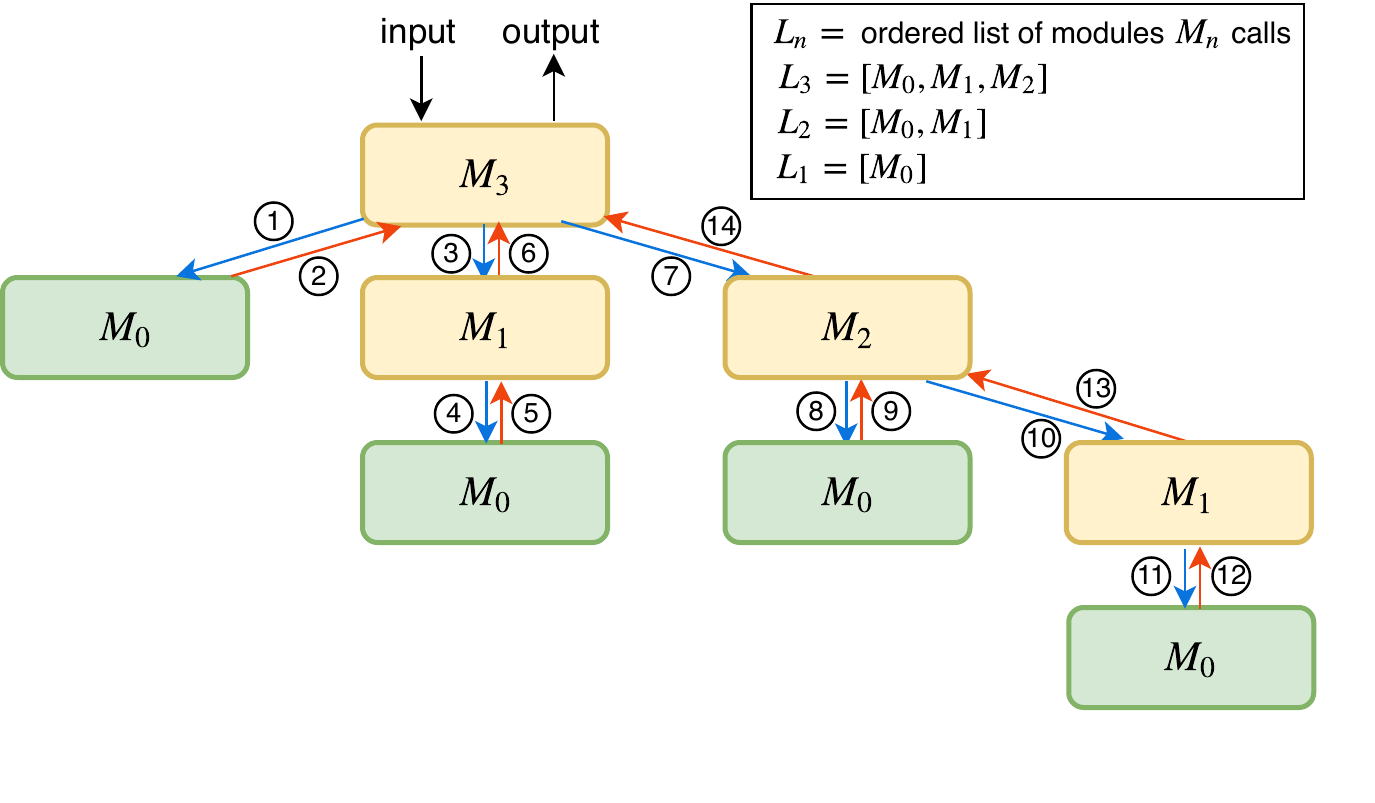}
\end{center}
\end{minipage}
\begin{minipage}{0.44\linewidth}
 %\vspace{1mm}
\caption{\footnotesize An example computation graph for \PMN~with four tasks.
Green rectangles denote terminal modules, and yellow rectangles denote compositional modules.
Blue arrows and red arrows represent calling and receiving outputs from submodules, respectively.
White numbered circles denote computation order.
For convenience, assume task levels correspond to the subscripts.
Calling $M_3$ invokes a chain of calls (blue arrows) to lower level modules which stop at the terminal modules.
%\SK{would it make sense to also show residual here? or too complicated?}
}
\label{fig:overview}
\end{minipage}
\vspace{-10mm}
\end{figure}

%\begin{figure}
%\vspace{-1.5mm}
%\begin{center}
%\includegraphics[width=0.44\linewidth,trim=0 70 0 0,clip=true]{high-level_pmnn1.pdf}
%\hspace{12mm}
%\includegraphics[width=0.39\linewidth,trim=150 218 335 190,clip=true]{reverse_tree_pmnn.pdf}
%\end{center}
%\vspace{-2.5mm}
%\caption{\footnotesize Overview of \PMN.
%Rectangles with a single border denote terminal modules and double borders denote compositional modules.
%Each arrow represents a communication.
%%For convenience, assume task levels correspond to the subscripts and all modules have access to environment $\E$.
%\textbf{Left}: General architecture of a compositional module.
%Given input and environment variables $\E$, it calls lower level modules, gathers information and produces an output.
%This diagram represents calling $M_k$ %in the sequence $\mathcal M_n=[M_1,...,M_k,...,M_{n-1}]$
%where red arrows indicate current activity.
%For clarity, all connections from $Q$ to other modules are not shown.
%\textbf{Right}: An example computation graph for \PMN~with four tasks.
%Note that $M_3$ does not call $M_1$ directly, as it need not use a lower level module unless necessary.}
%\label{fig:pmn}
%\vspace{-2.0mm}
%\end{figure}

\vspace{-2mm}
\section{Progressive Module Networks}
\label{sec:pmn}
\vspace{-1.5mm}

% Structure is important.
% The human brain has different parts for solving different problems.
% Moreover, each part has distinct patterns of information flow.
% Some are recurrent and some receive information from other parts which are routed differently.
% \SK{brain things, needs citation}
% Neural networks are regarded as black boxes since why and how they work are hard to interpret.
% Rather than fully connecting million of neurons and letting them figure out the optimal weights and information flow structure, it is sometimes better to have a structure that can aid the network's learning process.
% This is especially true in situations where data is scarce or learning an input-to-output mapping is difficult by simply stacking multiple layers.
% We give neural networks structure by designing modules to solve a specific task, and let them learn to communicate with each other.
% Analogous to a child's learning process, Progressive Module Networks (PMN) start from simple tasks and then progressively build on top of them to solve more complex tasks.
% \SF{is the children thing too strong? :)}
%\SK{i have commented out the part about brain and structure and so on. maybe move to/combine with intro, but seems too generic in model section.}

Most complex reasoning tasks can be broken down into a series of sequential reasoning steps.
We hypothesize that there exists a hierarchy with regards to complexity and order of execution:
high level tasks (\eg~counting) are more complex and benefit from leveraging outputs from lower level tasks (\eg~classification).
For any task, Progressive Module Networks (\PMN) learn a module that requests and uses outputs from lower modules to aid in solving the given task.
\textbf{This process is compositional, \ie, lower-level modules may call modules at an even lower level.}
%This process is \emph{compositional}, \ie, lower-level modules may call modules at an even lower level.
Solving a task corresponds to executing a directed acyclic computation graph where each node represents a module (see Fig.~\ref{fig:overview}).

PMN has a plug-and-play architecture where modules can be replaced by their improved versions.
This opens up promising ways to train intelligent robots that can compartmentalize and perform multiple tasks while progressively learning from experience and improving abilities.
PMN also chooses %\SK{chooses? sounds hard-gate. weights? doesn't sound so nice ;)}
which lower level modules to use through a soft-gating mechanism.
\textbf{A natural consequence of PMN's modularity and gating mechanism is interpretability.}
While we do not need to know the internal workings of modules, we can examine the queries and replies along with the information about which modules were used to reason about why the parent module produced a certain output.

Formally, given a task $n$ at level $i$, the task module $M_n$ can query other modules $M_k$ at level $j$ such that $j < i$.
Each module is designed to solve a particular task (output its best prediction) given an input and environment $\E$.
Note that $\E$ is accessible to every module and represents a broader set of ``sensory'' information available to the model.
For example, $\E$ may contain visual information such as an image, and text in the form of words (\ie, question).
\PMN~has two types of modules:
(i) \emph{terminal modules} execute the simplest tasks that do not require information from other modules (Sec.~\ref{sec:terminal_modules}); and
(ii) \emph{compositional modules} that learn to efficiently communicate and exploit lower-level modules to solve a task (Sec.~\ref{sec:compositional_modules}).
We describe the tasks studied in this paper in Sec.~\ref{sec:pmn_tasks} and provide a detailed example of how PMN is implemented and executed for VQA (Sec.~\ref{sec:example_vqa}).
% Alg.~\ref{alg:comp} summarizes the computation process of each module.

\vspace{-2.0mm}
\subsection{Terminal Modules}
\label{sec:terminal_modules}
\vspace{-1mm}

Terminal modules %do not query other modules, and
are by definition at the lowest level $0$.
%\SK{conflicts with Fig. 1, where lowest level and task is 1? Kinda}
They are analogous to base cases in a recursive function.
Given an input query $q$, a terminal module $M_\ell$ generates an output
%\begin{equation}
$o = M_\ell(q)$,
%\end{equation}
where $M_\ell$ is implemented with a neural network.
A typical example of a terminal module is an object classifier that takes as input a visual descriptor $q$, and predicts the object label $o$.

\vspace{-2mm}
\subsection{Compositional Modules}
\label{sec:compositional_modules}
\vspace{-1mm}

A compositional module $M_n$ makes a sequence of calls to lower level modules which in turn make calls to their children, in a manner similar to depth-first search (see Fig.~\ref{fig:overview}).
We denote the list of modules that $M_n$ is allowed to call by $\mathcal{L}_n=[M_m,\ldots,M_l]$.
Every module in $\mathcal L_n$ has level lower than $M_n$.
Since lower modules need not be sufficient in fully solving the new task, we optionally include a terminal module $\Delta_n$ that performs ``residual'' reasoning.
Also, many tasks require an attention mechanism to focus on certain parts of data.
We denote $\Omega_n$ as a terminal module that performs such soft-attention.
$\Delta_n$ and $\Omega_n$ are optionally inserted to the list $\mathcal L_n$ and treated as any other module.
%\SK{appended here would make people think added to the end. maybe change to inserted? included?}

The compositional aspect of \PMN~means that modules in $\mathcal L_n$ can have their own hierarchy of calls.
We make $\mathcal L_n$ an ordered list, where calls are being made in a sequential order, starting with the first in the list.
This way, information produced by earlier modules can be used when generating the query for the next.
For example, if one module is performing object detection, we may want to use its output (bounding box proposals), for querying other modules such as an attribute classifier.

For this work, the list $\mathcal L_n$, and thus the levels of tasks, are determined by hand.
Relaxing this and letting the model learn the task hierarchy itself is a challenging direction that we leave for future work.
Also, notice that the number of back-and-forth communications increases exponentially if each module makes use of every lower-level module.
Thus, in practice we restrict the list $\mathcal L_n$ to those lower-level modules that may intuitively  be needed by the task.
We emphasize that $M_n$ still (softly) chooses between them, and thus the expert intervention only removes the lower-level modules that are uninformative to the task.

\iffalse
To prevent this, we introduce some structural decisions.
While an intelligent machine should have the ability to choose which module to use, a machine learning practitioner with relevant domain knowledge can help make this choice.
For instance, a module that predicts the sentiment of a sentence would most likely not be useful for a object classifier module that predicts object labels given image regions.
Thus, we design modules and reduce the number of computations by
(i) restricting the list of relevant lower modules $L_n=[M_m,\ldots,M_l]$; and
(ii) allowing the parent module to use a subset among $L_n$.
Conditioned on the input, $M_n^i$ may need to use different lower level modules.
We assume every module in $L_n$ for $M_n^i$ has level lower than $i$ (by definition) and omit the superscript for brevity.

\SF{this paragraph needs to be shortened}
This process is analogous to how one writes a software program.
The \texttt{main()} function calls other functions with parameters (that in turn call their own), and uses returned information without needing to consider how exactly the called functions work internally.
It also has a sequence of commands with conditional and iterative statements.
Without loss of generality, $L_n$ can represent the order of execution steps.
Thus, any lower level module can appear any number of times in $L_n$.
Designing an order of execution, helps reduces the burden of learning greatly.
\fi

Our compositional module $M_n$ runs (pre-determined) $T_n$ passes over the list $\mathcal L_n$.
It keeps track of a state variable $s^t$ at time step $t \le T_n$.
%Our compositional module $M_n$ keeps track of a state variable $s^t$ at time step $t$.
This contains useful information obtained by querying other modules.
For example, $s^t$ can be the hidden state of a Recurrent Neural Network.
Each time step corresponds to executing \emph{every} module in $\mathcal L_n$ and updating the state variable. %, and modules have pre-specified number of steps $T_n$.
We describe the module components below, and Algorithm~\ref{alg:comp} shows how the computation is performed.
An example implementation of the components and demonstration of how they are used is detailed in Sec.~\ref{sec:example_vqa}.

%For the VQA task, we choose $s^t$ to be a tuple $(q_\mathrm{vqa}^t, u^{t-1})$ where $q_\mathrm{vqa}^t$ is a question vector for the current time step $t$ and $u^{t-1}$ is cumulated information at time $t-1$.

\textbf{State initializer.}\hspace{0.5mm}
Given a query (input) $q_n$, the initial state $s^1$ is produced using a \emph{state initializer} $I_n$.
%It could be a simple MLP or an assignment function such as in our VQA implementation, $I_\mathrm{vqa}$ that sets $s_1=(q_\mathrm{vqa}^1,u^0)$ where $q_\mathrm{vqa}^1=q_\mathrm{vqa}$ and $u^0$ is a zero vector.
%Details in Appendix~\ref{sec:appendix_module_details}.

\textbf{Importance function.}\hspace{2mm}
For each module $M_k$ (and $\Delta_n$, $\Omega_n$) in $\mathcal L_n$, we compute an importance score $g_n^k$ with $G_n(s^t)$. %The purpose of $g_k^t$ is to weight the importance of the $k$-th module's output to the current computation. It allows $M_n$ to softly choose which modules to use.
The purpose of $g_n^k$ is to enable $M_n$ to (softly) choose which modules to use.
This also enables training all module components with backpropagation.
%This also enables end-to-end model training with backpropagation.
%\SK{end-to-end here refers to all the local functions of module, but not lower modules. this becomes clear later, but maybe now too: ``enables training all module components with backprop''?}
Notice that $g_n^k$ is input dependent, and thus the module $M_n$ can effectively control which lower-level module outputs to use in state $s^t$. % using $g_k^t$.
Here, $G_n$ can be implemented as an MLP followed by either a softmax over submodules, or a sigmoid that outputs a score for each submodule.
%Using $g_k^t$ as a soft gate allows us to train the model end-to-end with back-propagation.
However, note that the proposed setup can be modified to adopt hard-gating mechanism using a threshold or sampling with reinforcement learning. %ignore lower modules that are deemed unimportant
%(\eg~using a threshold or sampling, and adopting RL).

\textbf{Query transmitter and receiver.}\hspace{2mm}
A query for module $M_k$ in $\mathcal L_n$ is produced using a \emph{query transmitter}, as $q_k = Q_{n\rightarrow k}(s^t, V, G_n(s^t))$.
%\SK{replace $G_n(s^t)$ by $g_n^k$? or does it use all $g_n$?} - sometime multiple scores are used
The output $o_k=M_k (q_k)$ received from $M_k$ is modified using a \emph{receiver function}, as $v_k = R_{k\rightarrow n}(s^t, o_k)$.
One can think of these functions as translators of the inputs and outputs into the module's own ``language".
Note that each module has a scratch pad $V$ to store outputs it receives from a list of lower modules $\mathcal L_n$, \ie, $v_k$ is stored to $V$.

%Note that the residual module $\Delta_n$ is called directly, \ie, we do not use the query transmitter and receiver.

\textbf{State update function.}\hspace{2mm}
After every module in $\mathcal L_n$ is executed, module $M_n$ updates its internal state using a \emph{state update function} $U_n$ as $s^{t+1} = U_n(s^t, V, \E, G_n(s^t))$.
This completes one time step of the module's computation.
Once the state is updated, the scratch pad $V$ is wiped clean and is ready for new outputs.
An example can be a simple gated sum of all outputs, \ie,~$s^{t+1} = \sum_k g_n^k\cdot v_k$.
%For a more complicated example, $U_\mathrm{vqa}$ passes $u^t = \sum g_n^k\cdot v_k$ as an input to a GRU (whose initial state is initialized with $q_\mathrm{vqa}^1$) that produces a query for the next time step, $q_\mathrm{vqa}^{t+1}$.
%Then, $s_{t+1}$ is set to $(q_\mathrm{vqa}^{t+1}, u^t)$.

\textbf{Prediction function.}\hspace{2mm}
After $T_n$ steps, the final module output is produced using a \emph{prediction function} $\Psi_n$ as $o_n = \Psi_n(s^1, \ldots, s^{T_n}, q_n, \E)$.
%As an example, $\Psi_n$ can be a mean-pool over MLP on $(s^t,q_n)$ for all $t$, or directly an MLP on $s_{T_n}$.

All module functions: state initializer $I$, importance function $G$, query transmitter $Q$, receiver $R$, state update function $U$, residual module $\Delta$, attention module $\Omega$, and prediction function $\Psi$ are implemented as neural networks or simple assignment functions (\eg set $q_k = v_l$).
%Note that all variables (\eg $o_k, q_k, v_k, s^t$) are vectorized so that gradient flow is easily possible.
Note that all variables (\eg $o_k, q_k, v_k, s^t$) are continuous vectors to allow learning with standard backpropagation.
For example, the output of the relationship detection module that predicts an object bounding box is a $N$ dimensional soft-maxed vector (assuming there are total of $N$ boxes or image regions in $\E$).

\textbf{Training.}\hspace{2mm}
We train our modules sequentially, from low level to high level tasks, one at a time.
The internal weights of the lower level modules are not updated, thus preserving their performance on the original task.
The new module only learns to communicate with them via the query transmitter $Q$ and receiver $R$.
We do train the weights of $\Delta$ and $\Omega$.
We train $I$, $G$, $Q$, $R$, $U$, and $\Psi$, by allowing gradients to pass through the lower level modules.
The loss function depends on the task $n$.

\begin{center}
\vspace{-2mm}
\algtext*{EndFunction}
\algtext*{EndFor}
\algtext*{EndIf}
\begin{algorithm}[t!]
\caption{\small Computation performed by our Progressive Module Network, for one module $M_n$}
\begin{footnotesize}
\begin{algorithmic}[1]
\Function{$M_n$}{$q_n$}                               \Comment{$\E$ and $\mathcal L_n$ are global variables}
  \State $s^1$ = $I_n(q_n)$                           \Comment{initialize the state variable}
  \For{$t \gets 1\,$ to $\, T_n$}                     \Comment{$T_n$ is the maximum time step}
    \State $V = []$                                   \Comment{wipe out scratch pad $V$}
    \State $g_n^1,\ldots,g_n^{|\mathcal L_n|} = G_n(s^t)$ \Comment{compute importance scores}
    \For{$k \gets 1$ to $|\mathcal L_n|$}             \Comment{$\mathcal L_n$ is the sequence of lower modules $[M_m,...,M_l]$}
      \State $q_k = Q_{n\rightarrow k}(s^t, V, G_n(s^t))$                  \Comment{produce query for $M_k$}
      \State $o_k = \mathcal L_n[k](q_k)$                    \Comment{call $k^{th}$ module $M_k = \mathcal L_n[k]$, generate output}
      \State $v_k = R_{k\rightarrow n}(s^t, o_k)$                       \Comment{receive and project output}
      \State $V.\mathrm{append}(v_k)$     \Comment{write $v_k$ to pad $V$}
    \EndFor
    \State $s^{t+1} = U_n(s^{t}, V, \E, G_n(s^t))$              \Comment{update module state}
  \EndFor
  \State $o_n$ = \Call{$\Psi_n$}{$s^1, \ldots, s^{T_n}, q_n, \E $} \Comment{produce the output}
  \State \Return $o_n$
\EndFunction
\end{algorithmic}
\end{footnotesize}
\label{alg:comp}
\end{algorithm}
\vspace{-2mm}
\end{center}

%!TEX root = paper.tex

\subsection{Progressive Module Networks for Visual Reasoning}
\label{sec:pmn_tasks}

We present an example of how \PMN~can be adopted for several tasks related to visual reasoning.
In particular, we consider six tasks: object classification, attribute classification, relationship detection, object counting, image captioning, and visual question answering.
%The level at which we consider each task is written as a superscript in the following sections.
Our environment $\E$ consists of:
{\bf (i)} \emph{image regions}: $N$ image features $X=[X_1, \ldots, X_N]$, each $X_i \in \mathbb{R}^{d}$ with corresponding bounding box coordinates $\mathbf{b}=[b_1, \ldots, b_N]$ extracted from Faster R-CNN~\citep{ren15}; and
{\bf (ii)} \emph{language}: vector representation of a sentence $S$ (in our example, a question).
$S$ is computed through a Gated Recurrent Unit~\citep{cho14} by feeding in word embeddings $[w_1,\ldots,w_T]$ at each time step.
%\SK{call (i) objects as image objects? or image regions? i think calling it objects makes it sound like you know what they are. maybe even ``image object proposals''?}

%Note that when we say `attend' or `attention map', we refer to soft-attention mechanism over the $N$ image regions.
%input, output %and the sequence of lower modules $\mathcal L$.
Below, we discuss each task and a module designed to solve it.
We provide detailed implementation and execution process of the VQA module in Sec.~\ref{sec:example_vqa}.
For other modules, we present a brief overview of what each module does in this section.
Further implementation details of all module architectures are in Appendix~\ref{sec:appendix_module_details}.

\textbf{Object and Attribute Classification (level 0).}\hspace{2mm}
Object classification is concerned with naming the object that appears in the image region, while attribute classification predicts the object's attributes (\eg~color).
% Object and attribute classification require naming the object that appears in the image region, or providing the object's attributes (\eg~color).
As these two tasks are fairly simple (not necessarily easy), we place $\Mobj$ and $\Matt$ as terminal modules at level $0$.
%The object classification module answers what is the object in a given image region, while attribute classifier predicts properties (\eg~color) of the region.
$\Mobj$ consists of an MLP that takes as input a visual descriptor for a bounding box $b_i$, \ie, $q_\mathrm{obj} = X_i$, and produces
% The object classification module consists of an MLP and produces
%\begin{equation}
$o_\mathrm{obj} = \Mobj (q_\mathrm{obj})$,
%\end{equation}
the penultimate vector prior to classification.
Attribute module $\Matt$ has a similar structure. %, but different parameters.
%We include an additional linear layer (classifier) operating on module outputs for projecting them to class logits only to train the module.
These are the only modules for which we do not use actual output labels, as we obtained better results for higher level tasks empirically.
%The models are trained to minimize the cross-entropy loss for predicting object/attribute classes.

\textbf{Image Captioning (level 1).}\hspace{2mm}
In image captioning, one needs to produce a natural language description of the image.
We design our module $\Mcap$ as a compositional module that uses information from
% $\Omega_\mathrm{cap}$, $\Mobj$, $\Matt$, and $\Delta_\mathrm{cap}$, \ie,
$\mathcal L_\mathrm{cap} = [\Omega_\mathrm{cap}, \Mobj, \Matt, \Delta_\mathrm{cap}]$.
We implement the state update function as a two-layer GRU network with $s^t$ corresponding to the hidden states.
Similar to~\citet{anderson17}, at each time step, the attention module $\Omega_\mathrm{cap}$ attends over image regions $X$ using the hidden state of the first layer.
The attention map $m$ is added to the scratch pad $V$.
The query transmitters produce a query (image vector at the attended location) using $m$ to obtain  nouns $\Mobj$ and adjectives $\Matt$.
The residual module $\Delta_\mathrm{cap}$ processes other image-related semantic information.
%\SK{and non-visual sentence filling words?} - sentence filling is done by the state updater
The outputs from modules in $\mathcal L_\mathrm{cap}$ are projected to a common vector space (same dimensions) by the receivers and stored in the scratch pad.
Based on their importance score, the gated sum of the outputs is used to update the state.
The natural language sentence $o_{cap}$ is obtained by predicting a word at each time step using a fully connected layer on the hidden state of the second GRU layer.

\textbf{Relationship Detection (level 1).}\hspace{2mm}
In this task the model is expected to produce triplets in the form of ``subject - relationship - object''~\citep{lu2016visual}.
We re-purpose this task as one that involves finding the relevant item (region) in an image that is related to a given input through a given relationship.
%\SK{do you want to clarify that this can be subject if given obj or obj if given subj}
The input to the module is $q_\mathrm{rel}=[b_i, r]$ where $b_i$ is a one-hot encoding of the input box %(with $i$-th entry is 1 and others 0)
and $r$ is a one-hot encoding of the  relationship category (\eg~above, behind).
The module produces $o_\mathrm{rel}=b_\mathrm{out}$ corresponding to the box for the subject/object related to the input $b_i$ through $r$.
We place $\Mrel$ on the first level as it may use object and attribute information that can be useful to infer relationships, \ie, $\mathcal L_\mathrm{rel} = [\Mobj, \Matt, \Delta_\mathrm{rel}]$.
We train the module using the cross-entropy loss.
\begin{figure}[t!]
\vspace{-6mm}
\begin{center}
\includegraphics[width=0.95\linewidth,trim=0 0 0 0]{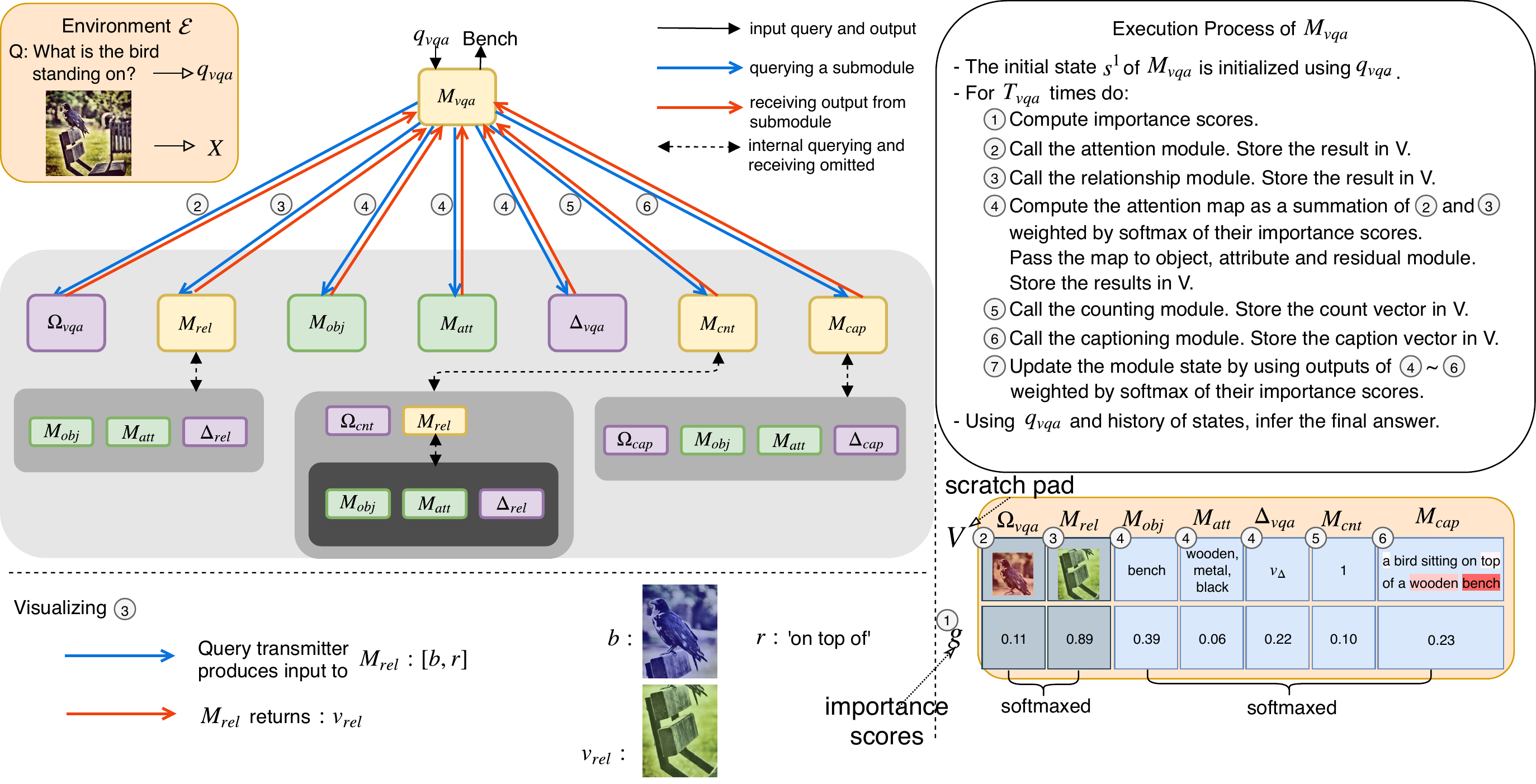}
\end{center}
\vspace{-2mm}
\caption{\small Example of \PMN's module execution trace on the VQA task.
Numbers in circles indicate the order of execution.
Intensity of gray blocks represents depth of module calls.
All variables including queries and outputs stored in $V$ are continuous vectors to allow learning with standard backpropagation
(\eg, caption is composed of a sequence of softmaxed $W$ dimensional vectors for vocabulary size $W$).
For $\Mcap$, words with higher intensity in red are deemed more relevant by $R_\mathrm{vqa}^\mathrm{cap}$.
%The yellow, green and purple boxes denote compositional, terminal, and residual/attention modules, respectively.
\textbf{Top:} high level view of module execution process.
\textbf{Bottom right:} computed importance scores and populated scratch pad.
Note that we perform the first softmax operation on $(\Omega_\mathrm{vqa}, \Mrel)$ to obtain an attention map and the second on $(\Mobj$, $\Matt$, $\Delta_\mathrm{vqa}$, $\Mcnt$, $\Mcap)$ to obtain the answer.
\textbf{Bottom left:} visualizing the query $\Mvqa$ sends to $\Mrel$, and the received output.}
\label{fig:execution_vqa}
\vspace{-4mm}
\end{figure}

\textbf{Object Counting (level 2).}\hspace{2mm}
Our next task is counting the number of objects in the image.
Given a vector representation of a natural language question (\eg~how many cats are in this image?), the goal of this module is to produce a numerical count.
The counting task is at a higher-level since it may also require us to understand relationships between objects.
For example, ``how many cats are on the blue chair?'', requires counting cats \emph{on top of} the blue chair.
We thus place $\Mcnt$ on the second level and provide it access to $\mathcal L_\mathrm{cnt} = [\Omega_\mathrm{cnt}, \Mrel]$.
The attention module $\Omega_\mathrm{cnt}$ finds relevant objects by using the input question vector.
$\Mcnt$ may also query $\Mrel$ if the question requires relational reasoning.
To answer ``how many cats are on the blue chair'', we can expect the query transmitter $Q_{\mathrm{cnt}\rightarrow \mathrm{rel}}$ to produce a query $q_\mathrm{rel} = [b_i, r]$ for the relationship module $\Mrel$ that includes the chair bounding box $b_i$ and relationship ``on top of'' $r$ so that $\Mrel$ outputs boxes that contain cats on the chair.
Note that both $\Omega_\mathrm{cnt}$ and $\Mrel$ produce attention maps on the boxes.
The state update function softly chooses a useful attention map by calculating softmax on the importance scores of $\Omega_\mathrm{cnt}$ and $\Mrel$.
For prediction function $\Psi_\mathrm{cnt}$, we adopt the counting algorithm by ~\cite{zhang18}, which builds a graph representation from attention maps to count objects.
$\Mcnt$ returns $o_\mathrm{cnt}$ which is the count vector corresponding to softmaxed one-hot encoding of the count (with maximum count $\in \mathbb{Z}$). %(representation of the number).

\textbf{Visual Question Answering (level 3).}\hspace{2mm}
VQA is our final and most complex task.
Given a vector representation of a natural language question, $q_\mathrm{vqa}$, the VQA module $\Mvqa$ uses $\mathcal L_\mathrm{vqa} = [\Omega_\mathrm{vqa}, \Mrel, \Mobj, \Matt, \Delta_\mathrm{vqa}, \Mcnt, \Mcap]$.
Similar to $\Mcnt$, $\Mvqa$ makes use of $\Omega_\mathrm{vqa}$ and $\Mrel$ to get an attention map.
The produced attention map is fed to the downstream modules $[\Mobj, \Matt, \Delta_\mathrm{vqa}]$ using the query transmitters.
$\Mvqa$ also queries $\Mcnt$ which produces a count vector.
For the last entry $\Mcap$ in $\mathcal L_\mathrm{vqa}$, the receiver attends over the words of the entire caption produced by $\Mcap$ to find relevant answers.
% $\Mcap$ produces a caption for the whole image, the receiver attends over the words in the produced caption to find relevant answers.
The received outputs are used depending on the importance scores.
Finally, $\Psi_\mathrm{vqa}$ produces an output vector based on $q_\mathrm{vqa}$ and all states $s^t$.
%\SK{This seems like a very clean explanation :) Good stuff! My one concern is that this does not necessarily follow the intuition that the reader may have about $\mathcal L$ having modules of decreasing complexity. Although that is also not the definition any more.}

\subsection{Example: $\Mvqa$ for Visual Question Answering}
\label{sec:example_vqa}
\vspace{-1mm}

We give a detailed example of how PMN is implemented for the VQA task.
The entire execution process is depicted in Fig.~\ref{fig:execution_vqa}, and the general algorithm is tabulated in Alg.~\ref{alg:comp}.
%\SK{in figure:
%1. $\Mcnt$ is shown as $\Mrel$ and then omega. Explanation above is [omega, $\Mrel$].
%2. $\Mcap$ is missing omega-cap.
%3. $\Mrel$ in the bottom should return $o_{rel}$? attention maps are o in description}

The input $q_\mathrm{vqa}$ is a vector representing a natural language question (\ie. the sentence vector $S \in \E$).
The state variable $s^t$ is represented by a tuple $(q_{\mathrm{vqa}}^t, k^{t-1})$ where $q_{\mathrm{vqa}}^t$ represents query to ask at time $t$ and $k^{t-1}$ represents knowledge gathered at time $t-1$.
The state initializer $I_\mathrm{vqa}$ is composed of a GRU with hidden state dimension 512.
The first input to GRU is $q_\mathrm{vqa}$, and $I_\mathrm{vqa}$ sets $s^1 = (q_\mathrm{vqa}^1, \textbf{0})$ where $q_\mathrm{vqa}^1$ is the first hidden state of the GRU and $\textbf{0}$ is a zero vector (no knowledge at first).

For $t$ in $T_{\mathrm{vqa}}=2$, $M_\mathrm{vqa}$ does the following seven operations:
\vspace{-2mm}
\begin{enumerate}[\hspace{0pt}(1)]
\item{
The importance function $G_\mathrm{vqa}$ is executed. It is implemented as a linear layer $\mathbb{R}^{512} \to \mathbb{R}^7$ (for the seven modules in $\mathcal L_\mathrm{vqa}$) that takes $s^t$, specifically $q_\mathrm{vqa}^t \in s^t$ as input.
}

\item{
$Q_{\mathrm{vqa}\rightarrow \mathrm{\Omega}}$ passes $q_{\mathrm{vqa}}^t$ to the attention module $\Omega_\mathrm{vqa}$ which attends over the image regions $X$ with $q_{\mathrm{vqa}}^t$ as the key vector.
$\Omega_\mathrm{vqa}$ is implemented as an MLP that computes a dot-product soft-attention similar to ~\citet{yang16san}. The returned attention map $v_\mathrm{\Omega}$ is added to the scratch pad $V$.
}

\item{
$Q_{\mathrm{vqa}\rightarrow \mathrm{rel}}$ produces an input tuple $[b, r]$ for $\Mrel$.
The input object box $b$ is produced by a MLP that does soft attention on image boxes, and the relationship category $r$ is produced through a MLP with $q_{\mathrm{vqa}}^t$ as input.
$\Mrel$ is called with $[b, r]$ and the returned map $v_\mathrm{rel}$ is added to $V$.
}
%\SK{hm.. $o_{rel}$? Either is ok, but let's stick to same everywhere. In that case, need to change $m_\Omega$ to $o_\Omega$ too}

\item{
$Q_{\mathrm{vqa}\rightarrow\mathrm{obj}}$, $Q_{\mathrm{vqa}\rightarrow\mathrm{att}}$, and $Q_{\mathrm{vqa}\rightarrow\mathrm{\Delta}}$ first compute a joint attention map $m$ as summation of $(v_\mathrm{\Omega}, v_\mathrm{rel})$ weighted by the softmaxed importance scores of $(\Omega_\mathrm{vqa}, \Mrel)$,
and they pass the sum of visual features $X$ weighted by $m$ to the corresponding modules.
$\Delta_\mathrm{vqa}$ is implemented as an MLP.
The receivers project the outputs into 512 dimensional vectors $v_\mathrm{obj}$, $v_\mathrm{att}$, and $v_\mathrm{\Delta}$ through a sequence of linear layers, batch norm, and $\mathrm{tanh}()$ nonlinearities. They are added to $V$.
%The receivers project $o_\mathrm{obj}$, $o_\mathrm{att}$, and $\delta_\mathrm{vqa}$ into 512 dimensional vectors $v_\mathrm{obj}$, $v_\mathrm{att}$, and $v_\mathrm{\Delta}$ through a sequence of linear layers, batch norm, and $\mathrm{tanh}()$ nonlinearities. They are added to $V$.
}

\item{
$Q_{\mathrm{vqa}\rightarrow \mathrm{cnt}}$ passes $q_\mathrm{vqa}^t$ to $\Mcnt$ which returns $o_\mathrm{cnt}$.
$R_{\mathrm{cnt}\rightarrow\mathrm{vqa}}$ projects the count vector $o_\mathrm{cnt}$ into a 512 dimensional vector $v_{cnt}$ through the same sequence of layers as above.
$v_\mathrm{cnt}$ is added to $V$.
}

\item{
$\Mvqa$ calls  $\Mcap$ and
% $Q_{\mathrm{vqa}\rightarrow\mathrm{cap}}$ calls $\Mcap$.
$R_{\mathrm{cap}\rightarrow\mathrm{vqa}}$ receives natural language caption of the image.
It converts words in the caption into vectors $[w_1,\ldots,w_T]$ through an embedding layer.
The embedding layer is initialized with 300 dimensional GloVe vectors~\citep{pennington14} and fine-tuned.
It does softmax attention operation over $[w_1,\ldots,w_T]$ through a MLP with $q_{\mathrm{vqa}}^t \in s_t$ as the key vector, resulting in word probabilities $p_1,\ldots,p_T$.
The sentence representation $\sum_{i}^{T} p_i \cdot w_i$ is projected into a 512 dimensional vector $v_\mathrm{cap}$ using the same sequence as $v_\mathrm{cnt}$. $v_\mathrm{cap}$ is added to $V$.
}

\item{
The state update function $U_\mathrm{vqa}$ first does softmax operation over the importance scores of $(\Mobj$, $\Matt$, $\Delta_\mathrm{vqa}$, $\Mcnt$, $\Mcap)$.
We define an intermediate knowledge vector $k^t$ as the summation of $(v_\mathrm{obj}, v_\mathrm{att}, \delta_\mathrm{vqa}, v_\mathrm{cnt}, v_\mathrm{cap})$ weighted by the softmaxed importance scores.
$U_\mathrm{vqa}$ passes $k^t$ as input to the GRU initialized by $I_\mathrm{vqa}$, and we get $q_\mathrm{vqa}^{t+1}$ the new hidden state of the GRU.
The new state $s^{t+1}$ is set to $(q_\mathrm{vqa}^{t+1}, k^t)$.
This process allows the GRU to compute new question and state vectors based on what has been \emph{asked} and \emph{seen}.
}
\end{enumerate}

After $T_{\mathrm{vqa}}$ steps, the prediction function $\Psi_\mathrm{vqa}$ computes the final output based on the initial question vector $q_\mathrm{vqa}$ and all knowledge vectors $k^t \in s^t$.
Here, $q_\mathrm{vqa}$ and $k^t$ are fused with gated-tanh layers and fed through a final classification layer similar to ~\citet{anderson17}, and the logits for all time steps are added.
The resulting logit is the final output $o_\mathrm{vqa}$ that corresponds to an answer in the vocabulary of the VQA task.
Note that the exact form of each module can be different.
While we leave a more general architecture across tasks as future work, we stress that one of PMN's strengths is that once a module is trained, it can be used as a blackbox by the higher-level modules.
Details of other modules' architectures are provided in Appendix~\ref{sec:appendix_module_details}.

%!TEX root = paper.tex
\vspace{-2mm}
\section{Experiments}
\label{sec:experiments}
\vspace{-1mm}

We present experiments demonstrating the impact of progressive learning of modules.
We also analyze and evaluate the reasoning process of PMN as it is naturally interpretable. 
We conduct experiments on three datasets (see Appendix~\ref{subsec:appendix_datasets} for details): Visual Genome (VG)~\citep{krishna16}, VQA 2.0~\citep{goyal17}, MS-COCO~\citep{lin14}.
These datasets contain natural images and are thus more complex in visual appearance and language diversity than CLEVR~\citep{johnson2017clevr} that contains synthetic scenes.
Neural module networks~\citep{andreas16,hu17} show excellent performance on CLEVR but their performance on natural images is quite below the state-of-the-art.
For all datasets, we extract bounding boxes and their feature representations using a pretrained model from~\citet{anderson17}.% which is a Faster-RCNN~\citep{ren15} based on ResNet-101~\citep{he16}.

\vspace{-2mm}
\subsection{Progressive Learning of Tasks and Modules}
\vspace{-1mm}

\textbf{Object and Attribute Classification.}\hspace{2mm}
We train these modules with annotated bounding boxes from the VG dataset.
We follow ~\citet{anderson17} and use 1,600 and 400 most commonly occurring object and attribute classes, respectively.
Each extracted box is associated with the ground truth label of the object with greatest overlap.
It is ignored if there are no ground truth boxes with IoU $>$ 0.5.
This way, each box is annotated with one object label and zero or more attribute labels.
$M_\mathrm{obj}$ achieves 54.9\% top-1 accuracy and 86.1\% top-5 accuracy. We report mean average precision (mAP) for attribute classification which is a multi-label classification problem. $M_\mathrm{att}$ achieves 0.14 mAP and 0.49 weighted mAP.
\emph{mAP} is defined as the mean over all classes, and \emph{weighted mAP} is weighted by the number of instances for each class.
As there are a lot of redundant classes (\eg~car, cars, vehicle) and boxes have sparse attribute annotations, the accuracy may seem artificially low.

\textbf{Image Captioning.}\hspace{2mm}
We report results on MS-COCO for image captioning.
We use the standard split from the 2014 captioning challenge to avoid data contamination with VQA 2.0 or VG.
This split contains 30\% less data compared to that proposed in~\citet{karpathy15} that most current works adopt.
We report performance using the CIDEr~\citep{vedantam15} metric.
A baseline (non-compositional module) achieves a strong CIDEr score of 108.
Using the object and attribute modules we are able to obtain 109 CIDEr.
While this is not a large improvement, we suspect a reason for this is the limited vocabulary.
The MS-COCO dataset has a fixed set of 80 object categories and does not benefit by using knowledge from modules that are trained on more diverse data.
We believe the benefits of  \PMN~would be clearer on a diverse captioning dataset with many more object classes.
Also, including high-level modules such as $\Mvqa$ would be an interesting direction for future work.

% \begin{wraptable}{R}{0.45\linewidth}
% % \begin{table}[H]
% \begin{small}
% \begin{center}
% \tabcolsep=0.12cm
% \vspace{-0.5cm}
% \begin{tabular}{ccccc}
% \toprule
% \multirow{2}{*}{\textbf{Model}} & \multicolumn{3}{c}{\textbf{Composition}} & \multirow{2}{*}{\textbf{CIDEr}} \\
%  & BASE & OBJ & ATT & \\
% \midrule
% $\rMcap{0}$ & \cmark &   -     &    -    & 108 \\
% $\rMcap{1}$ & \cmark & $\Mobj$ & $\Matt$ & 109 \\
% \bottomrule
% \end{tabular}
% \caption{\small Performance of $\Mcap$ with and without helper modules $\Mobj, \Matt$.}
% \label{tbl:cap_table}
% \vspace{-0.4cm}
% \end{center}
% \end{small}
% % \end{table}
% \end{wraptable}

% We report performance using the CIDEr~\cite{vedantam15} metric in Table~\ref{tbl:cap_table}, and achieve a strong baseline score of 108. The baseline here refers to a non-compositional module.

\begin{wraptable}{R}{0.35\linewidth}
\begin{small}
\vspace{-5mm}
\caption{\small Performance of $\Mrel$} % w/without $\Mobj$ and $\Matt$.}
%In subject-relationship-object tuple, object acc. is for locating objects, and subject acc. for subjects.}
\vspace{-2mm}
\label{tbl:rel_table}
\begin{center}
\tabcolsep=0.1cm
\vspace{-2.5mm}
\scalebox{0.8}{
\begin{tabular}{cccccc}
\toprule
\multirow{2}{*}{\textbf{Model}} & \multicolumn{3}{c}{\textbf{Composition}} & \multicolumn{2}{c}{\textbf{Acc. (\%)}} \\
 & BASE & OBJ & ATT & Object & Subject \\
\midrule
%Random      &   -    &   -     &    -    & 2.8 & 2.8 \\
$\rMrel{0}$ & \cmark &   -     &    -    & 51.0 & 55.9  \\
$\rMrel{1}$ & \cmark & $\Mobj$ & $\Matt$ & 53.4 & 57.8  \\
\bottomrule
\end{tabular}
}

\end{center}

\end{small}

\begin{small}
\caption{\small Accuracy for $\Mcnt$}
%Using the relationship module in addition to objects and attributes boosts performance by a significant amount.}
\label{tbl:cnt_table}
\vspace{-3mm}
\begin{center}
\tabcolsep=0.1cm
\scalebox{0.8}{
\begin{tabular}{cccccc}
\toprule
\multirow{2}{*}{\textbf{Model}} & \multicolumn{4}{c}{\textbf{Composition}} & \multirow{2}{*}{\textbf{Acc. (\%)}} \\
 & BASE & OBJ & ATT & REL & \\
\midrule
%Random      &   -    &   -     &    -    &     -       & {\color{red}add} \\
$\rMcnt{0}$ & \cmark &   -     &    -    &     -       & 45.4  \\
$\rMcnt{1}$ & \cmark & $\Mobj$ & $\Matt$ &     -       & 47.4  \\
$\rMcnt{2}$ & \cmark & $\Mobj$ & $\Matt$ & $\rMrel{1}$ & 50.0  \\
\bottomrule
\end{tabular}
}

\vspace{-5mm}
\end{center}
\end{small}
\end{wraptable}

\textbf{Relationship Detection.}\hspace{2mm}
We use top 20 commonly occurring relationship categories, which are defined by a set of words with similar meaning (\eg~in, inside, standing in).
Relationship tuples in the form of ``subject - relationship - object'' are extracted from Visual Genome~\citep{krishna16,lu2016visual}.
We train and validate the relationship detection module using 200K/38K train/val tuples that have both subject and object boxes overlapping with the ground truth boxes (IoU $>$ 0.7).
Table~\ref{tbl:rel_table} shows improvement in performance when using modules.
Even though accuracy is relatively low, model errors are reasonable, qualitatively.
This is partially attributed to multiple correct answers although there is only one ground truth answer.

\textbf{Object Counting.}\hspace{2mm}
We extract questions starting with `how many' from VQA 2.0 which results in a training set of $\sim$50K questions.
We additionally create $\sim$89K synthetic questions based on the VG dataset by counting the object boxes and forming `how many' questions.
This synthetic data helps to increase the accuracy by $\sim$1\% for all module variants. % which shows ablation studies for the counting module.
Since the number of questions that have relational reasoning and counting (\eg~how many people are sitting on the sofa? how many plates on table?) is limited, we also sample relational synthetic questions from VG.
These questions are used only to improve the parameters of query transmitter $Q_\mathrm{cnt\rightarrow rel}$ for the relationship module.
% See Appendix A for details.
Table~\ref{tbl:cnt_table} shows a large improvement (4.6\%) of the compositional module over the non-modular baseline.
When training for the next task (VQA), unlike other modules whose parameters are fixed, we \emph{fine-tune} the counting module because counting module expects the same form of input - embedding of natural language question.
The performance of counting module depends crucially on the quality of attention map over bounding boxes.
By employing more questions from the whole VQA dataset, we obtain a better attention map, and the performance of counting module increases from 50.0\% (\cf~Table~\ref{tbl:cnt_table}) to 55.8\% with finetuning
(see Appx~\ref{sec:appendix_training} for more details).

\begin{table}[t!]
\vspace{-2mm}
\begin{minipage}{0.35\linewidth}
\caption{\small Model ablation for VQA.
We report mean$\pm$std computed over three runs.
Steady increase indicates that information from modules helps, and that \PMN~makes use of lower modules effectively.
The base model $\rMvqa{0}$ does not use any lower level modules other than the residual and attention modules. %, or equivalently is predicted by the residual function.
}
\label{tbl:vqa_table}
\end{minipage}
\begin{minipage}{0.65\linewidth}
\vspace{-3mm}
% \vspace{1mm}

\begin{center}
\begin{scriptsize}
\addtolength{\tabcolsep}{-2.0pt}
\begin{tabular}{cccccccc}
\toprule
\multirow{2}{*}{\textbf{Model}} & \multicolumn{6}{c}{\textbf{Composition}} & \multirow{2}{*}{\textbf{Accuracy (\%)}} \\
 & BASE & OBJ & ATT & REL & CNT & CAP & \\
\midrule
$\rMvqa{0}$ & \cmark &    -    &    -    &      -      &      -      &     -       & 62.05 {\scriptsize $\pm$0.11} \\
$\rMvqa{1}$ & \cmark & $\Mobj$ & $\Matt$ &      -      &      -      &     -       & 63.38 {\scriptsize $\pm$0.05}  \\
$\rMvqa{2}$ & \cmark & $\Mobj$ & $\Matt$ & $\rMrel{1}$ &      -      &     -       & 63.64 {\scriptsize $\pm$0.07}  \\
$\rMvqa{3}$ & \cmark & $\Mobj$ & $\Matt$ &      -      & $\rMcnt{1}$ &     -       & 64.06 {\scriptsize $\pm$0.05}  \\
$\rMvqa{4}$ & \cmark & $\Mobj$ & $\Matt$ & $\rMrel{1}$ & $\rMcnt{2}$ &     -       & 64.36 {\scriptsize $\pm$0.06}  \\
$\rMvqa{5}$ & \cmark & $\Mobj$ & $\Matt$ & $\rMrel{1}$ & $\rMcnt{2}$ & $\rMcap{1}$ & 64.68 {\scriptsize $\pm$0.04}  \\
\bottomrule
\end{tabular}
\end{scriptsize}
\end{center}

\end{minipage}
% \vspace{-2mm}
\end{table}

\begin{table}[t!]
\begin{center}
\vspace{-2mm}
\caption{\small Comparing VQA accuracy of \PMN$\,$ with state-of-the-art models.
Rows with Ens \cmark denote ensemble models.
test-dev is development test set and test-std is standard test set for VQA 2.0.}
\vspace{-3mm}
\tabcolsep=0.14cm
\begin{small}
\scalebox{0.8}{
\begin{tabular}{lc|cccc|cccc|cccc}
\toprule
\multirow{2}{*}{\textbf{Model}} & \multirow{2}{*}{\textbf{Ens}} & \multicolumn{4}{c|}{\textbf{VQA 2.0 val}} & \multicolumn{4}{c|}{\textbf{VQA 2.0 test-dev}} & \multicolumn{4}{c}{\textbf{VQA 2.0 test-std}} \\
 &  & Yes/No & Number & Other & All & Yes/No & Number & Other & All & Yes/No & Number & Other & All   \\
\midrule
~\cite{andreas16}    & -       & 73.38 & 33.23 & 39.93 & 51.62    &  -     & -     & -     & -    & -     & -     & -     & -    \\
~\cite{yang16san}       & -       & 68.89 & 34.55 & 43.80 & 52.20    &  -     & -     & -     & -    & -     & -     & -     & -     \\
~\cite{teney17}         & -       & 80.07 & 42.87 & 55.81 & 63.15    & 81.82 & 44.21 & 56.05 & 65.32 & 82.20 & 43.90 & 56.26 & 65.67 \\
~\cite{teney17}         & \cmark  & -     & -     & -     & -        & 86.08 & 48.99 & 60.80 & 69.87 & 86.60 & 48.64 & 61.15 & 70.34 \\
~\cite{zhou18}          & -       & -     & -     & -     & -        & 84.27 & 49.56 & 59.89 & 68.76 & -     & -     & -     & - \\
~\cite{zhou18}          & \cmark  & -     & -     & -     & -        & -     & -     & -     & -     & 86.65 & 51.13 & 61.75 & 70.92 \\
~\cite{zhang18}         & -       & -     & 49.36 & -     & 65.42    & 83.14 & 51.62 & 58.97 & 68.09 & 83.56 & 51.39 & 59.11 & 68.41 \\
~\cite{kim2018bilinear}*    & -       & - & - & - & 66.04    &  85.43     & 54.04     & 60.52     & 70.04    & 85.82     & 53.71     & 60.69    & 70.35    \\
~\cite{kim2018bilinear}*        & \cmark       & -     & - & -     & -    & 86.68 & 54.94 & 62.08 & 71.40 & 87.22 & 54.37 & 62.45 & 71.84 \\
~\cite{jiang2018pythia}*         & \cmark       & -     & - & -     & -    & 87.82 & 51.54 & 63.41 & 72.12 & 87.82 & 51.59 & 63.43 & 72.25 \\

\midrule
baseline $\rMvqa{0}$                      & -       & 80.28 & 43.06 & 53.21 & 62.05    &  -     & -     & -     & -    & -     & -     & -     & -  \\
\PMN~$\rMvqa{5}$                          & -       & 82.48 & 48.15 & 55.53 & 64.68    & 84.07 & 52.12 & 57.99 & 68.07 & -     & -     & -     & -  \\
\PMN~$\rMvqa{5}$                          & \cmark  & -     & -     & -     & -        & 85.74 & 54.39 & 60.60 & 70.25 & 86.34 & 54.26 & 60.80 & 70.68 \\
\bottomrule
\end{tabular}
}
\end{small}

\label{tbl:vqa_compare}
\end{center}
\vspace{-3.5mm}
\end{table}

\textbf{Visual Question Answering.}\hspace{2mm}
We present ablation studies on the val set of VQA 2.0 in Table~\ref{tbl:vqa_table}.
As seen, \PMN~strongly benefits from utilizing different modules, and achieves a performance improvement of 2.6\% over the baseline.
Note that all results here are without additional questions from the VG data.
We also compare performance of \PMN~for the VQA task with state-of-the-art models in Table~\ref{tbl:vqa_compare}.
%\SK{Following tradition},
Models are trained on the train split for results on VQA val, while for test-dev and test-std, models are trained on both the train and val splits.
Although we start with a much lower baseline performance of 62.05\% on the val set (vs. 65.42\%~\citep{zhang18}, 63.15\%~\citep{teney17}, 66.04\%~\citep{kim2018bilinear}), \PMN's performance is on par with these models.
Note that entries with * are parallel works to ours.
Also, as~\cite{jiang2018pythia} showed, the performance depends strongly on engineering choices such as learning rate scheduling, data augmentation, and ensembling models with different architectures.

\textbf{Three additional experiments on VQA.}\hspace{2mm}
(1) To verify that the gain is not merely from the increased model capacity, we trained a baseline model with the number of parameters approximately matching the total number of parameters of the full \PMN~model.
This baseline with more capacity also achieves 62.0\%, thus confirming our claim.
(2) We also evaluated the impact of the additional data available to us.
We convert the subj-obj-rel triplets used for the relationship detection task to additional QAs (e.g. Q: what is on top of the desk?, A: laptop) and train the $\rMvqa{1}$ model (Table~\ref{tbl:vqa_table}).
This results in an accuracy of 63.05\%, not only lower than $M_\mathrm{vqa_2}$ (63.64\%) that uses the relationship module via PMN, but also lower than $M_\mathrm{vqa_1}$ at 63.38\%.
This suggests that while additional data may change the question distribution and reduce performance, PMN is robust and benefits from a separate relationship module.
(3) Lastly, we conducted another experiment to show that PMN does make efficient use of the lower level modules.
We give equal importance scores to all modules in $M_\mathrm{vqa_5}$ model (Table~\ref{tbl:vqa_table}) (thus, fixed computation path), achieving 63.65\% accuracy.
While this is higher than the baseline at 62.05\%, it is lower than $M_\mathrm{vqa_5}$ at 64.68\% which softly chooses which sub-modules to use. %which modules are more important.

% \vspace{-2mm}
% \subsection{Comparison to State-of-the-Art}
% We compare performance of \PMN~for the VQA task with state-of-the-art models in Table~\ref{tbl:vqa_compare}.
% Although we start with a much lower baseline performance of 61.81\% on the val set (vs. 65.42\%~\cite{zhang18} and 63.15\%~\cite{teney17}), \PMN's performance is on par with these models.
% For results for VQA val, models are trained on the train split. For test-dev and test-std, models are trained on both the train and val splits.

\vspace{-2mm}
\subsection{Interpretability Analysis}
\vspace{-1mm}

%We now analyze the interpretability and reasoning that \PMN~affords.

\textbf{Visualizing the model's reasoning process.}
We present a qualitative analysis of the answering process.
%This allows us to peek into the reasoning process, and generated output.
In Fig.~\ref{fig:execution_vqa}, $\Mvqa$ makes query $q_\mathrm{rel}=[b_i, r]$ to $\Mrel$ where $b_i$ corresponds to the blue box `bird' and $r$ corresponds to `on top of' relationship.
$\Mvqa$ correctly chooses (\ie higher importance score) to use  $\Mrel$ rather than its own output produced by $\Omega_\mathrm{vqa}$ since the question requires relational reasoning.
With the attended green box obtained from $\Mrel$, $\Mvqa$ mostly uses the object and captioning modules to produce the final answer.
More examples are presented in Appendix ~\ref{sec:appendix_pmn_exec}.

\begin{wraptable}{R}{0.3\linewidth}

\begin{center}
\caption{\small Average human judgments from 0 to 4.
\cmark~indicates that model got final answer right, and \xmark~for wrong.}
%Note that our model is rated higher consistently, even when \PMN~and baseline are both correct or wrong.}
\label{tbl:vqa_mt}
\vspace{-3mm}
\tabcolsep=0.12cm

\begin{small}
\scalebox{0.77}{
\begin{tabular}{ccccc}
\toprule
\multicolumn{2}{c}{\textbf{Correct?}} & \multirow{2}{*}{\textbf{\# Q}} & \multicolumn{2}{c}{\textbf{Human Rating}} \\
\PMN   & Baseline & & \PMN & Baseline \\
\midrule
\cmark & \cmark & 715 & 3.13 & 2.86  \\
\cmark & \xmark & 584 & 2.78 & 1.40  \\
\xmark & \cmark & 162 & 1.73 & 2.47  \\
\xmark & \xmark & 139 & 1.95 & 1.66  \\
\midrule
\multicolumn{2}{c}{All images} & 1600 & 2.54 & 2.24 \\
\bottomrule
\end{tabular}
}
\end{small}
\vspace{-0.5cm}
\end{center}
\end{wraptable}

\begin{wrapfigure}{R}{0.4\textwidth}
\vspace{-3mm}
% \includegraphics[width=0.4\linewidth,trim=0 0 600 0,clip=true]{pmn_reasoning_fig3-H-crop.pdf}
% \hfill
% \includegraphics[width=0.4\linewidth,trim=600 0 0 0,clip=true]{pmn_reasoning_fig3-H-crop.pdf}
\includegraphics[width=\linewidth]{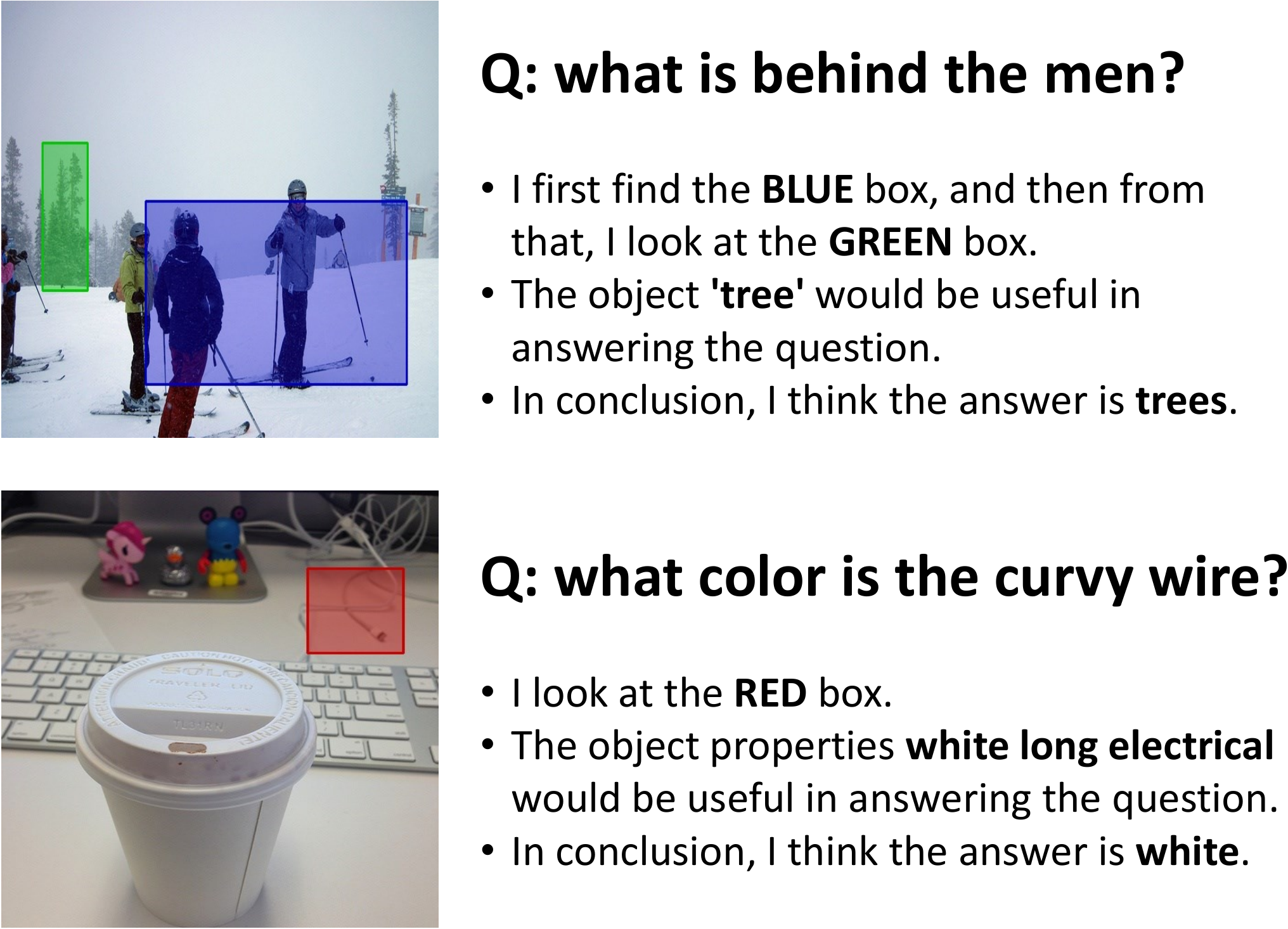}
\vspace{-5mm}
\caption{\small Example of  \PMN's reasoning processes.
\textbf{Top}: it correctly first find a person and then uses relationship module to find the tree behind him.
\textbf{Bottom}: it finds the wire and then use attribute module to correctly infer its attributes - white, long, electrical - and then outputs the correct answer. }
\label{fig:pmn_examples}
\vspace{-4mm}
\end{wrapfigure}

\textbf{Judging Answering Quality.}
The modular structure and gating mechanism of \PMN~makes it easy to interpret the reasoning behind the outputs.
We conduct a human evaluation with Amazon Mechanical Turk on 1,600 randomly chosen questions. % to measure the interpretability of the VQA module.
Each worker is asked to rate the explanation generated by the baseline model and the \PMN~like a teacher grades student exams.
The baseline explanation is composed of the bounding box it attends to and the final answer.
For \PMN, we form a rule-based natural language explanation based on the prominent modules used.
An example is shown in Fig.~\ref{fig:pmn_examples}.

% \begin{figure}[t!]
% \end{figure}

Each question is assessed by three human workers.
The workers are instructed to score how satisfactory the explanations are in the scale of 0 (very bad), 1 (bad), 2 (satisfactory), 3 (good), 4 (very good).
Incorrect reasoning steps are penalized, so if \PMN~produces wrong reasoning steps that do not lead to the correct answer, it could get a low score.
On the other hand, the baseline model often scores well on simple questions that do not need complex reasoning (\eg what color is the cat?).

We report results in Table~\ref{tbl:vqa_mt}, and show more examples in Appendix~\ref{sec:appendix_pmn_reasoning}.
Human evaluators tend to give low scores to wrong answers and high scores to correct answers regardless of explanations, but \PMN~always scores higher if both \PMN~and baseline gets a question correct or wrong.
Interestingly, a correct answer from \PMN~gets 1.38 points higher than wrong baseline, but a correct baseline scores only 0.74 higher than a wrong \PMN~answer.
This shows that \PMN~gets partial marks even when it gets an answer wrong since the reasoning steps are partially correct.

\textbf{Low Data Regime.}
\PMN~benefits from re-using modules and only needs to learn the communication between them.
This allows us to achieve good performance even when using a fraction of the training data.
Table~\ref{tbl:vqa_data} presents the absolute gain in accuracy \PMN~achieves.
For this experiment, we use $\mathcal L_\mathrm{vqa}=[\Omega_{vqa}, \Mrel, \Mobj, \Matt, \Delta_{vqa}, \Mcap]$ (because of overlapping questions from $\Mcnt$).
When the amount of data is really small (1\%), \PMN~does not help because there is not enough data to learn to communicate with lower modules.
The maximum gain is obtained when using 10\% of data.
It shows that \PMN~can help in situations where there is not a huge amount of training data since it can exploit previously learned knowledge from other modules.
The gain remains constant at about 2\% from then on.

\begin{table}[H]
\vspace{-1mm}
\begin{minipage}{0.4\linewidth}
\caption{\small Absolute gain in accuracy when using a fraction of the training data.}
\label{tbl:vqa_data}

\end{minipage}
\begin{minipage}{0.6\linewidth}
\tabcolsep=0.1cm
\begin{center}
\begin{scriptsize}
\begin{tabular}{lcccccc}
\toprule
\textbf{Fraction of VQA training data (in \%)} & 1 & 5 & 10 & 25 & 50  & 100  \\
\midrule
\textbf{Absolute accuracy gain (in \%)} &  -0.49 & 2.21 &  4.01 &  2.66 &  1.79 &  2.04   \\
\bottomrule
\end{tabular}
\end{scriptsize}
\end{center}
\end{minipage}
\vspace{-6mm}
\end{table}

\vspace{-2mm}
\section{Conclusion and Discussion}
\label{sec:conclusion}
\vspace{-1mm}

In this work, we proposed Progressive Module Networks (PMN) that train task modules in a compositional manner, by exploiting previously learned lower-level task modules.
PMN can produce queries to call other modules and make use of the returned information to solve the current task.
Given experts in specific tasks, the parent module only needs to learn how to effectively communicate with them.
It can also choose which lower level modules to use.
Thus, PMN is data efficient and provides a more interpretable reasoning processes.
Also, since there is no need to know about the inner workings of the children modules,
it opens up promising ways to train intelligent robots that can compartmentalize and perform multiple tasks as they progressively improve their abilities.
Moreover, one task can benefit from unrelated tasks unlike conventional multi-task learning algorithms.

%It is also an important step towards more intelligent machines as it can easily accommodate novel and increasingly more complex tasks.
PMN as it stands has few limitations with respect to hand-designed structures and the need for additional supervision.
Nevertheless, PMN is an important step towards more interpretable, compositional multi-task models.
Some of the questions to be solved in the future include:
1) learning module lists automatically;
2) choosing few modules (hard attention) to reduce overhead;
3) more generic structure of module components across tasks; and
4) joint training of all modules.

\textbf{Acknowledgments.}
Partially supported by the DARPA Explainable AI (XAI) program, Samsung, and NSERC.
We also thank NVIDIA for their donation of GPUs.

%%%%%%%%%%%%%%%%%%%%%%%%%%%%%%%%%%%%%%%%%%%%%%%%%%%%%%%%%%%%%%%%%%%%%%%%%%%%
%{\small
%\bibliographystyle{ieee}
%\bibliography{refs}
%}
\bibliography{refs}
\bibliographystyle{iclr2019_conference}

\newpage
\appendix
\section*{Appendices}
\addcontentsline{toc}{section}{Appendices}
\renewcommand{\thesection}{\Alph{section}}
%!TEX root = paper.tex
\section{Module Architectures}
\label{sec:appendix_module_details}
We discuss the detailed architecure of each module. We first describe the shared environment and soft attention mechanism architecture.

\textbf{Environment.}\hspace{2mm}
The sensory input that form our environment $\E$ consists of:
{\bf (i)} \emph{image regions}: $N$ image regions $X=[X_1, \ldots, X_N]$, each $X_i \in \mathbb{R}^{d}$ with corresponding bounding box coordinates $\mathbf{b}=[b_1, \ldots, b_N]$ extracted from Faster R-CNN~\citep{ren15}; and
{\bf (ii)} \emph{language}: vector representation of a sentence $S$ (in our example, a question).
$S$ is computed through a one layer GRU by feeding in the embedding of each word $[w_1,\ldots,w_T]$ at each time step.
For (i), we use a pretrained model from~\citet{anderson17} to extract features and bounding boxes.

\textbf{Soft attention.}\hspace{2mm}
For all parts that use soft-attention mechanism, an MLP is emloyed.
Given some \emph{key vector} $k$ and a set of data to be attended $\{d_1,\ldots,d_N\}$, we compute
\begin{equation}
\mathrm{attention\_map} = (z(f(k)\cdot g(d_1)), \ldots, z(f(k)\cdot g(d_N)))
\end{equation}
where $f$ and $g$ are a sequence of linear layer followed by ReLU activation function that project $k$ and $d_i$ into the same dimension, and $z$ is a linear layer that projects the joint representation into a single number.
Note that we do not specify softmax function here because sigmoid is used for some cases.

\subsection{Object and Attribute Classification (Level 0)}
The input to both modules $\Mobj, \Matt$ is a visual descriptor for a bounding box $b_i$ in the image, \ie,~$q_\mathrm{obj} = X_i$.
$\Mobj$ and $\Matt$ projects the visual feature $X_i$ to a 300-dimensional vector through a single layer neural network followed by $\mathrm{tanh}()$ non-linearity.
We expect this vector to represent the name and attributes of the box $b_i$.
%\SK{add a sentence like: We expect this vector to represent the word embedding of the class labels.
%Is the classifier layer made up of the GloVe embeddings? That ensures it, would be nice to say. This is also how a vector output from these modules can still be converted to a word easily.}
% - the embedding layer is not used anymore.

\subsection{Image Captioning (Level 1)}
$\Mcap$ takes zero vector as the model input and produces natural language sentence as the output based on the environment $\E$ (detected image regions in an image).
It has $\mathcal L_\mathrm{cap} = [\Omega_{\mathrm{cap}}, \Mobj, \Matt, \Delta_{\mathrm{cap}}]$ and goes through maximum of $T_\mathrm{cap}=16$ time steps or until it reaches the end of sentence token.
$\Mcap$ is implemented similarly to the captioning model in ~\citet{anderson17}.
We employ two layered GRU~\citep{cho14} as the recurrent state update function $U_\mathrm{cap}$ where $s^t = (h^t_1, h^t_2)$ with hidden states of the first and second layers of $U_\mathrm{cap}$. Each layer has 1000-d hidden states.

The state initializer $I_\mathrm{cap}$ sets the initial hidden state of $U_\mathrm{cap}$, or the model state $s^t$, as a zero vector.
For $t$ in $T_{\mathrm{cap}}=16$, $M_\mathrm{cap}$ does the following four operations:
\vspace{-2mm}
\begin{enumerate}[\hspace{0pt}(1)]
\item{
The importance function $G_\mathrm{cap}$ is executed. It is implemented as a linear layer $\mathbb{R}^{1000} \to \mathbb{R}^4$ (for the four modules in $\mathcal L_\mathrm{cap}$) that takes $s^t$, specifically $h_1^t \in s^t$ as input.
}

\item{
$Q_{\mathrm{cap}\rightarrow \mathrm{\Omega}}$ passes $h_1^t$ to the attention module $\Omega_\mathrm{cap}$ which attends over the image regions $X$ with $h_1^t$ as the key vector.
$\Omega_\mathrm{cap}$ is implemented as a soft-attention mechanism so that it produces attention probabilities $p_i$ (via softmax) for each image feature $X_i \in \E$.
The returned attention map $v_\mathrm{\Omega}$ is added to the scratch pad $V$.
}

\item{
$Q_{\mathrm{cap}\rightarrow\mathrm{obj}}$ and $Q_{\mathrm{cap}\rightarrow\mathrm{att}}$ pass the sum of visual features $X$ weighted by $v_\mathrm{\Omega} \in V$ to the corresponding modules.
$\Delta_\mathrm{cap}$ is implemented as an MLP.
The receivers project the outputs into 1000 dimensional vectors $v_\mathrm{obj}$, $v_\mathrm{att}$, and $v_\mathrm{\Delta}$ through a sequence of linear layers, batch norm, and $\mathrm{tanh}()$ nonlinearities.
They are added to $V$.
}

\item{
As stated above, $U_\mathrm{cap}$ is a two-layered GRU.
At time $t$, the first layer takes input the average visual features from the environment $\E$, $\frac{1}{N}\sum_{i}X_i$, embedding vector of previous word $w_{t-1}$, and $h^t_2$.
For time $t=1$, \emph{beginning-of-sentence} embedding and zero vector are inputs for $w_{1}$ and $h^{1}_1$, respectively.
The second layer is fed $h^t_1$ as well as the information from other modules,
\begin{equation}
\rho = \sum (\mathrm{softmax}(g_\mathrm{obj}, g_\mathrm{att}, g_{\Delta}) \cdot (v_\mathrm{obj}, v_\mathrm{att}, v_\mathrm{\Delta}))
\end{equation}
which is a gated summation of outputs in $V$ with softmaxed importance scores.
We now have a new state $s^{t+1} = (h^{t+1}_1, h^{t+1}_2)$.
}
\end{enumerate}

The output of $\Mcap$, $o_\mathrm{cap}$, is a sequence of words produced through $\Psi_\mathrm{cap}$ which is a linear layer projecting each $h^t_2$ in $s^t$ to the output word vocabulary.

\subsection{Relationship Detection (Level 1)}
Relationship detection task requires one to produce triplets in the form of ``subject - relationship - object''~\citep{lu2016visual}.
We re-purpose this task as one that involves finding the relevant item (region) in an image that is related to a given input through a given relationship.
The input to the module is $q_\mathrm{rel}=[b_i, r]$ where $b_i$ is a one-hot encoded input bounding box (whose $i$-th entry is 1 and others 0) and $r$ is a one-hot encoded relationship category (\eg~above, behind).
$\Mrel$ has $\mathcal L_\mathrm{rel} = [\Mobj, \Matt, \Delta_{\mathrm{rel}}]$ and goes through $T_{rel}=N$ steps where $N$ is the number of bounding boxes (image regions in the environment).
So at time step $t$, the module looks at the $t$-th box.
$\Mrel$ uses $\Mobj$ and $\Matt$ just as feature extractors for each bounding box.
Therefore, it does not have a complex structure.

The state initializer $I_\mathrm{rel}$ projects $r$ to a 512 dimensional vector with an embedding layer, and the resulting vector is set as the first state $s^1$.

For $t$ in $T_{\mathrm{rel}}=N$, $M_\mathrm{rel}$ does the following three operations:

\begin{enumerate}[\hspace{0pt}(1)]

\item{
$Q_{\mathrm{rel}\rightarrow\mathrm{obj}}$ and $Q_{\mathrm{rel}\rightarrow\mathrm{att}}$ pass the image vector corresponding to the bounding box $b_t$ to $\Mobj$ and $\Matt$.
$R_{\mathrm{obj}\rightarrow\mathrm{rel}}$ and $R_{\mathrm{att}\rightarrow\mathrm{rel}}$ are identity functions, \ie, we do not modify the object and attribute vectors.
The outputs $v_\mathrm{obj}$ and $v_\mathrm{att}$ are added to $V$.
}

\item{
$\Delta_\mathrm{rel}$ projects the coordinate of the current box $b_t$ to a 512 dimensional vector.
This resulting $v_\mathrm{\Delta}$ is added to $V$.
}

\item{
$U_\mathrm{rel}$ concatenates the visual feature $X_t$ with $v_\mathrm{obj}, v_\mathrm{att}, v_\mathrm{\Delta}$ from $V$.
The concatenated vector is fed through a MLP resulting in 512 dimensional vector.
This corresponds to the new state $s^{t+1}$.
}
\end{enumerate}
After $N$ steps, the prediction function $\Psi_\mathrm{rel}$ does the following operations:
\\
The first state $s^1$ which contains relationship information is multiplied element-wise with $s^{i+1}$ (Note: $s^{i+1}$ corresponds to the input box $b_i$).
Let such a vector be $l$.
It produces an attention map $b_\mathrm{out}$ over all bounding boxes in $b$.
The inputs to the attention function are $s^2,\ldots,s^{T_\mathrm{rel}}$ (i.e. all image regions) and the key vector $l$.
$o_\mathrm{rel} = b_\mathrm{out}$ is the output of $\Mrel$ which represents an attention map indicating the bounding box that contains the related object.

\subsection{Counting (Level 2)}
Given a vector representation of a natural language question (\eg~how many cats are in this image?), the goal of this module is to produce a count.
The input $q_\mathrm{cnt} = S \in \E$ is a vector representing a natural language question.
When training $\Mcnt$, $q_\mathrm{cnt}$ is computed through a one layer GRU with hidden size of 512 dimensions.
The input to the GRU at each time step is the embedding of each word from the question.
Word embeddings are initialized with 300 dimensional GloVe word vectors~\citep{pennington14} and fine-tuned thereafter.
Similar to visual features obtained through CNN, the question vector is treated as an environment variable.
$\Mcnt$ has $\mathcal L_\mathrm{cnt} = [\Omega_\mathrm{cnt}, \Mrel]$ and goes through only one time step.

The state initializer $I_\mathrm{cnt}$ is a simple function that just sets $s^1 = q_\mathrm{cnt}$.

For $t$ in $T_{\mathrm{cnt}}=1$, $M_\mathrm{cnt}$ does the following four operations:

\begin{enumerate}[\hspace{0pt}(1)]
\item{
The importance function $G_\mathrm{cnt}$ is executed. It is implemented as a linear layer $\mathbb{R}^{512} \to \mathbb{R}^2$ (for the two modules in $\mathcal L_\mathrm{cnt}$) that takes $s^t$ as input.
}

\item{
$Q_{\mathrm{cnt}\rightarrow \mathrm{\Omega}}$ passes $s^t$ to the attention module $\Omega_\mathrm{cnt}$ which attends over the image regions $X$ with $s^t$ as the key vector.
$\Omega_\mathrm{cnt}$ is implemented as an MLP that computes a dot-product soft-attention similar to ~\citet{yang16san}. The returned attention map $v_\mathrm{\Omega}$ is added to the scratch pad $V$.
}

\item{
$Q_{\mathrm{cnt}\rightarrow \mathrm{rel}}$ produces an input tuple $[b, r]$ for $\Mrel$.
The input object box $b$ is produced by a MLP that does soft attention on image boxes, and the relationship category $r$ is produced through a MLP with $s^t$ as input.
$\Mrel$ is called with $[b, r]$ and the returned map $v_\mathrm{rel}$ is added to $V$.
}

\item{
$U_\mathrm{cnt}$ first computes probabilities of using $v_\mathrm{\Omega}$ or $v_\mathrm{rel}$ by doing a softmax over the importance scores.
$v_\mathrm{\Omega}$ and $v_\mathrm{rel}$ are weighted and summed with the softmax probabilities resulting in the new state $s^2$ containing the attention map.
Thus, the state update function chooses the map from $\Mrel$ if the given question involves in relational reasoning.
}

\end{enumerate}

The prediction function $\Psi_\mathrm{cnt}$ returns a count vector.
The count vector is computed through the counting algorithm by ~\citet{zhang18}, which builds a graph representation from attention maps to count objects.
The method uses $s^2$ through a sigmoid and bounding box coordinates $b$ as inputs.
~\citet{zhang18} is a fully differentiable algorithm and the resulting count vector corresponds to one-hot encoding of a number.
We let the range of count be 0 to 12 $\in \mathbb{Z}$.
Please refer to~\citet{zhang18} for details of the counting algorithm.

\subsection{Visual Question Answering (Level 3)}
The description for the VQA task (Sec.~\ref{sec:example_vqa}) is included here again for completeness.
The input $q_\mathrm{vqa}$ is a vector representing a natural language question (\ie. the sentence vector $S \in \E$).
The state variable $s^t$ is represented by a tuple $(q_{\mathrm{vqa}}^t, k^{t-1})$ where $q_{\mathrm{vqa}}^t$ represents query to ask at time $t$ and $k^{t-1}$ represents knowledge gathered at time $t-1$.
The state initializer $I_\mathrm{vqa}$ is composed of a GRU with hidden state dimension 512.
The first input to GRU is $q_\mathrm{vqa}$, and $I_\mathrm{vqa}$ sets $s^1 = (q_\mathrm{vqa}^1, \textbf{0})$ where $q_\mathrm{vqa}^1$ is the first hidden state of the GRU and $\textbf{0}$ is a zero vector (no knowledge at first).

For $t$ in $T_{\mathrm{vqa}}=2$, $M_\mathrm{vqa}$ does the following seven operations:
\vspace{-2mm}
\begin{enumerate}[\hspace{0pt}(1)]
\item{
The importance function $G_\mathrm{vqa}$ is executed. It is implemented as a linear layer $\mathbb{R}^{512} \to \mathbb{R}^7$ (for the seven modules in $\mathcal L_\mathrm{vqa}$) that takes $s^t$, specifically $q_\mathrm{vqa}^t \in s^t$ as input.
}

\item{
$Q_{\mathrm{vqa}\rightarrow \mathrm{\Omega}}$ passes $q_{\mathrm{vqa}}^t$ to the attention module $\Omega_\mathrm{vqa}$ which attends over the image regions $X$ with $q_{\mathrm{vqa}}^t$ as the key vector.
$\Omega_\mathrm{vqa}$ is implemented as an MLP that computes a dot-product soft-attention similar to ~\citet{yang16san}. The returned attention map $v_\mathrm{\Omega}$ is added to the scratch pad $V$.
}

\item{
$Q_{\mathrm{vqa}\rightarrow \mathrm{rel}}$ produces an input tuple $[b, r]$ for $\Mrel$.
The input object box $b$ is produced by a MLP that does soft attention on image boxes, and the relationship category $r$ is produced through a MLP with $q_{\mathrm{vqa}}^t$ as input.
$\Mrel$ is called with $[b, r]$ and the returned map $v_\mathrm{rel}$ is added to $V$.
}
%\SK{hm.. $o_{rel}$? Either is ok, but let's stick to same everywhere. In that case, need to change $m_\Omega$ to $o_\Omega$ too}

\item{
$Q_{\mathrm{vqa}\rightarrow\mathrm{obj}}$, $Q_{\mathrm{vqa}\rightarrow\mathrm{att}}$, and $Q_{\mathrm{vqa}\rightarrow\mathrm{\Delta}}$ first compute a joint attention map $m$ as summation of $(v_\mathrm{\Omega}, v_\mathrm{rel})$ weighted by the softmaxed importance scores of $(\Omega_\mathrm{vqa}, \Mrel)$,
and they pass the sum of visual features $X$ weighted by $m$ to the corresponding modules.
$\Delta_\mathrm{vqa}$ is implemented as an MLP.
The receivers project the outputs into 512 dimensional vectors $v_\mathrm{obj}$, $v_\mathrm{att}$, and $v_\mathrm{\Delta}$ through a sequence of linear layers, batch norm, and $\mathrm{tanh}()$ nonlinearities. They are added to $V$.
%The receivers project $o_\mathrm{obj}$, $o_\mathrm{att}$, and $\delta_\mathrm{vqa}$ into 512 dimensional vectors $v_\mathrm{obj}$, $v_\mathrm{att}$, and $v_\mathrm{\Delta}$ through a sequence of linear layers, batch norm, and $\mathrm{tanh}()$ nonlinearities. They are added to $V$.
}

\item{
$Q_{\mathrm{vqa}\rightarrow \mathrm{cnt}}$ passes $q_\mathrm{vqa}^t$ to $\Mcnt$ which returns $o_\mathrm{cnt}$.
$R_{\mathrm{cnt}\rightarrow\mathrm{vqa}}$ projects the count vector $o_\mathrm{cnt}$ into a 512 dimensional vector $v_{cnt}$ through the same sequence of layers as above.
$v_\mathrm{cnt}$ is added to $V$.
}

\item{
$\Mvqa$ calls  $\Mcap$ and
% $Q_{\mathrm{vqa}\rightarrow\mathrm{cap}}$ calls $\Mcap$.
$R_{\mathrm{cap}\rightarrow\mathrm{vqa}}$ receives natural language caption of the image.
It converts words in the caption into vectors $[w_1,\ldots,w_T]$ through an embedding layer.
The embedding layer is initialized with 300 dimensional GloVe vectors~\citep{pennington14} and fine-tuned.
It does softmax attention operation over $[w_1,\ldots,w_T]$ through a MLP with $q_{\mathrm{vqa}}^t \in s_t$ as the key vector, resulting in word probabilities $p_1,\ldots,p_T$.
The sentence representation $\sum_{i}^{T} p_i \cdot w_i$ is projected into a 512 dimensional vector $v_\mathrm{cap}$ using the same sequence as $v_\mathrm{cnt}$. $v_\mathrm{cap}$ is added to $V$.
}

\item{
The state update function $U_\mathrm{vqa}$ first does softmax operation over the importance scores of $(\Mobj$, $\Matt$, $\Delta_\mathrm{vqa}$, $\Mcnt$, $\Mcap)$.
We define an intermediate knowledge vector $k^t$ as the summation of $(v_\mathrm{obj}, v_\mathrm{att}, \delta_\mathrm{vqa}, v_\mathrm{cnt}, v_\mathrm{cap})$ weighted by the softmaxed importance scores.
$U_\mathrm{vqa}$ passes $k^t$ as input to the GRU initialized by $I_\mathrm{vqa}$, and we get $q_\mathrm{vqa}^{t+1}$ the new hidden state of the GRU.
The new state $s^{t+1}$ is set to $(q_\mathrm{vqa}^{t+1}, k^t)$.
This process allows the GRU to compute new question and state vectors based on what has been \emph{asked} and \emph{seen}.
}
\end{enumerate}

After $T_{\mathrm{vqa}}$ steps, the prediction function $\Psi_\mathrm{vqa}$ computes the final output based on the initial question vector $q_\mathrm{vqa}$ and all knowledge vectors $k^t \in s^t$.
Here, $q_\mathrm{vqa}$ and $k^t$ are fused with gated-tanh layers and fed through a final classification layer similar to ~\citet{anderson17}, and the logits for all time steps are added.
The resulting logit is the final output $o_\mathrm{vqa}$ that corresponds to an answer in the vocabulary of the VQA task.

\newpage
\section{Additional Experimental Details}
In this section, we provide more details about datasets and module training.
% We also give more examples of execution traces of PMN on the visual question answering task.
\label{sec:appendix_exp_detail}

\subsection{Datasets}
\label{subsec:appendix_datasets}
We extract bounding boxes and their visual representations using a pretrained model from~\citet{anderson17}which is a Faster-RCNN~\citep{ren15} based on ResNet-101~\citep{he16}.
It produces 10 to 100 boxes with 2048-d feature vectors for each region.
To accelerate training, we remove overlapping bounding boxes that are most likely duplicates (area overlap IoU > 0.7) and keep only the 36 most confident boxes (when available).

\textbf{MS-COCO}
contains $\sim$100K images with annotated bounding boxes and captions.
It is a widely used dataset used  for benchmarking several vision tasks such as captioning and object detection.
%, and many annotations are based on these images.

\textbf{Visual Genome}
is collected to relate image concepts to image regions.
It has over 108K images with annotated bounding boxes containing 1.7M visual question answering pairs, 3.8M object instances, 2.8M attributes and 1.5M relationships between two boxes.
Since the dataset contains MS-COCO images, we ensure that we do not train on any MS-COCO validation or test images.

\textbf{VQA 2.0}
is the most popular visual question-answering dataset, with 1M questions on 200K natural images.
Questions in this dataset require reasoning about objects, actions, attributes, spatial relations, counting, and other inferred properties; making it
%This makes VQA 2.0
an ideal dataset for our visual-reasoning PMN.

\subsection{Training}
\label{sec:appendix_training}
Here, we give training details of each module.
We train our modules sequentially, from low level to high level tasks, one at a time.
When training a higher level module, internal weights of the lower level modules are not updated, thus preserving their performance on the original task. We do train the weights of the residual module $\Delta$ and the attention module $\Omega$.
We train $I$, $G$, $Q$, $R$, $U$, %residue computer $\Delta$,
and $\Psi$, by allowing gradients to pass through the lower level modules.
Thus, while the existing lower modules are held fixed, the new module learns to communicate with them via the query transmitter $Q$ and receiver $R$.

\textbf{Object and attribute classification.} $\Mobj$ is trained to minimize the cross-entropy loss for predicting object class by including an additional linear layer on top of the module output.
$\Matt$ also include an additional linear layer on top of the module output, and is trained to minimize the binary cross-entropy loss for predicting attribute classes since one detected image region can contain zero or more attribute classes. We make use of 780K/195K train/val object instances paired with attributes from the Visual Genome dataset.
They are trained with the Adam optimizer at learning rate of 0.0005 with batch size 32 for 20 epochs.

\textbf{Image captioning.} $\Mcap$ is trained using cross-entropy loss at each time step (maximum likelihood).
Parameters are updated using the Adam optimizer at learning rate of 0.0005 with batch size 64 for 20 epochs.
We use the standard split of MS-COCO captioning dataset.

\textbf{Relationship detection.} $\Mrel$ is trained using cross-entropy loss on ``subject - relationship - object'' pairs with Adam optimizer with learning rate of 0.0005 with batch size 128 for 20 epochs. The pairs are extracted from the Visual Genome dataset that have both subject and object boxes overlapping with the ground truth boxes (IoU $>$ 0.7), resulting in 200K/38K train/val tuples.

\textbf{Counting.} $\Mcnt$ is trained using cross-entropy loss on questions starting with `how many' from the VQA 2.0 dataset. We use Adam optimizer with learning rate of 0.0001 with batch size 128 for 20 epochs.
As stated in the experiments section, we additionally create $\sim$89K synthetic questions to increase our training set by counting the object boxes and forming `how many' questions from the VG dataset (\eg~(Q: how many dogs are in this picture?, A:3) from an image containing three bounding boxes of dog).
We also sample relational synthetic questions from each training image from VG that are used to train only the module communication parameters when the relationship module is included.
We use the same 200K/38K split from the relationship detection task by concatenating `how many'+subject+relationship' or `how many'+relationship+object (\eg~how many plates on table?, how many behind door?).
The module communication parameters for $\Mrel$ in this case are $Q_{\mathrm{cnt}\rightarrow\mathrm{rel}}$ which compute a relationship category and the input image region to be passed to $\Mrel$.
To be clear, we supervise $q_\mathrm{rel}=[b_i,r]$ to be sent to $\Mrel$ by reducing cross entropy loss on $b_i$ and $r$.

\textbf{Visual Question Answering.} $\Mvqa$ is trained using binary cross-entropy loss on $o_\mathrm{vqa}$ with Adam optimizer with learning rate of 0.0005 with batch size 128 for 7 epochs. We empirically found binary cross-entropy loss to work better than cross-entropy which was also reported by ~\cite{anderson17}.
Unlike other modules whose parameters are fixed, we \emph{fine-tune} only the counting module because counting module expects the same form of input - embedding of natural language question.
The performance of counting module depends crucially on the quality of attention map over bounding boxes.
By employing more questions from the whole VQA dataset, we obtain a better attention map, and the performance of counting module increases from 50.0\% to 55.8\% with finetuning.
Since $\Mvqa$ and $\Mcnt$ exepect the same form of input, the weights of attention modules $\Omega_\mathrm{\{vqa,cnt\}}$ and query transmitters for the relationship module $Q_\mathrm{\{vqa,cnt\} \rightarrow rel}$ are shared.

\newpage
\section{PMN Execution Illustrated}
\label{sec:appendix_pmn_exec}
\begin{figure}[!htb]
\includegraphics[width=\linewidth]{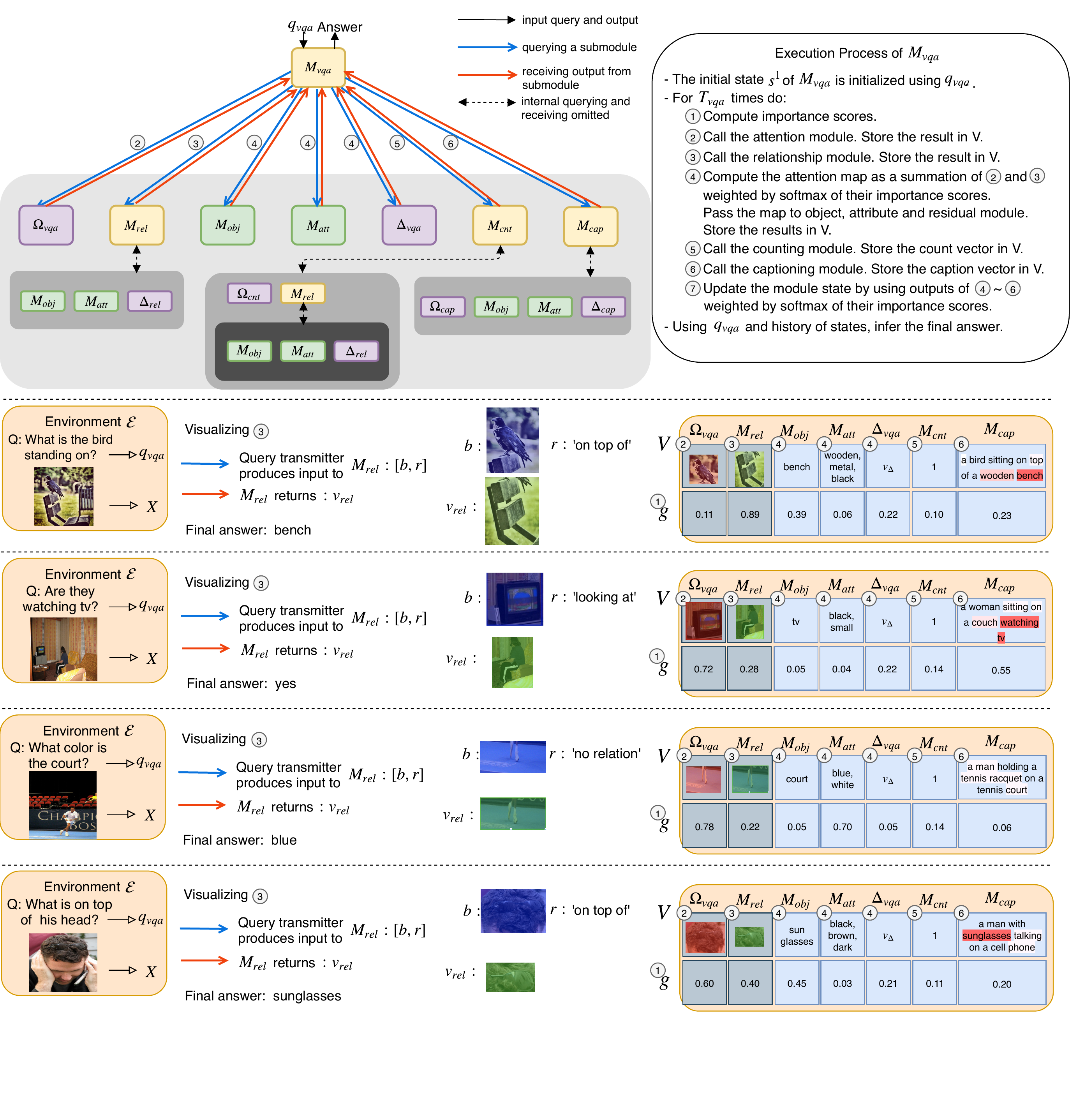}
\vspace{-15mm}
\caption{
\small Example of \PMN's module execution trace on the VQA task.
Numbers in circles indicate the order of execution.
Intensity of gray blocks represents depth of module calls.
All variables including queries and outputs stored in $V$ are vectorized to allow gradients to flow
(\eg, caption is composed of a sequence of softmaxed $W$ dimensional vectors for vocabulary size $W$).
For $\Mcap$, words with higher intensity in red are deemed more relevant by $R_\mathrm{vqa}^\mathrm{cap}$.
}
\label{fig:appendixb2}
\vspace{-2mm}
\end{figure}

We provide more examples of the execution traces of PMN on the visual question answering task in Figure~\ref{fig:appendixb2}.
Each row in the figure corresponds to different examples. For each row in the figure, the left column shows the environment $\E$, the middle column shows the final answer \& visualizes step 3 in the execution process, and
the right column shows computed importance scores along with populated scratch pad.

\newpage
\section{Examples of PMN's Reasoning}
\label{sec:appendix_pmn_reasoning}
\begin{figure}[!htb]
\vspace{-2mm}
\includegraphics[width=\linewidth,trim=0 0 0 0]{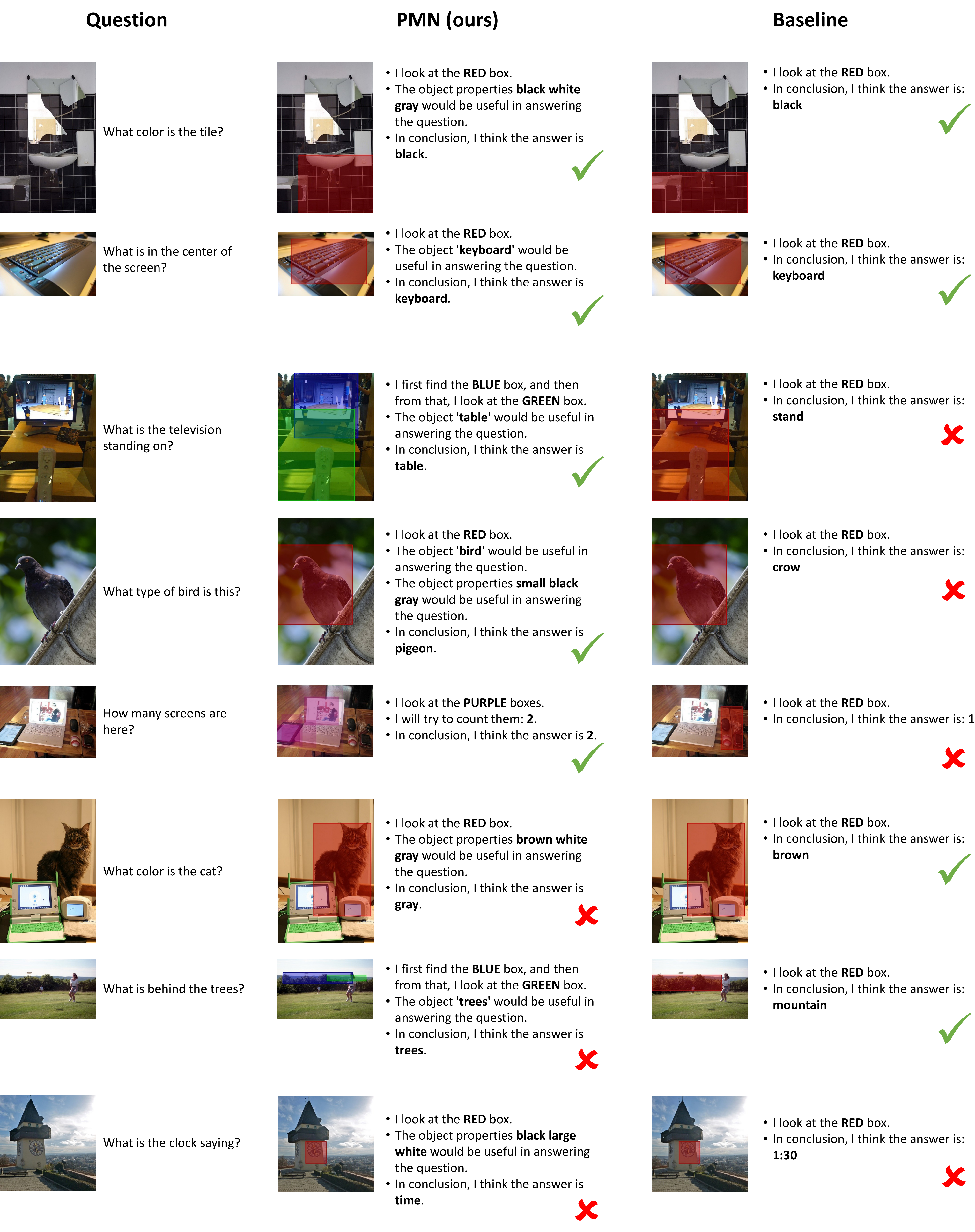}
\caption{\small Example of PMN's reasoning processes compared with the baseline given the question on the left. \cmark and \xmark~denote correct and wrong answers, respectively.}
\label{fig:appendixc}
\vspace{-2mm}
\end{figure}

We provide more examples of the human evaluation experiment on interpretability of PMN compared with the baseline model in Figure~\ref{fig:appendixc}.

\end{document}